\documentclass[final,5p,times,twocolumn]{elsarticle}

\usepackage{amssymb}
\usepackage{stmaryrd}
\usepackage{amsmath}
\usepackage{amsfonts}

\SetSymbolFont{stmry}{bold}{U}{stmry}{m}{n}

\usepackage{ifthen}
\usepackage{algorithm}
\usepackage{algorithmic}

\newcommand{\mv}{\boldsymbol{m}}
\newcommand{\cov}{\boldsymbol{C}}

\newcommand{\x}{\boldsymbol{x}}
\newcommand{\y}{\boldsymbol{y}}
\newcommand{\z}{\boldsymbol{z}}
\newcommand{\0}{\boldsymbol{0}}

\newcommand{\p}{\boldsymbol{p}}
\newcommand{\mueff}{\mu_{\mathrm{eff}}}
\newcommand{\vv}{\boldsymbol{v}}
\newcommand{\vsnr}{\boldsymbol{v}_{\mathrm{snr}}}
\newcommand{\vsnri}[1][i]{\boldsymbol{v}_{\mathrm{snr}, #1}}

\newcommand{\vgain}{\xi_{\mathrm{gain}}}
\newcommand{\vthresh}{\xi_{\mathrm{thresh}}}

\newcommand{\accs}{\boldsymbol{s}}
\newcommand{\accgamma}{\boldsymbol{\gamma}}
\newcommand{\Neff}{\hat{N}_{\mathrm{eff}}}
\newcommand{\Ntrueeff}{N_{\mathrm{eff}}}

\newcommand{\E}{\mathbb{E}}

\newcommand{\Var}{\mathrm{Var}}

\newcommand{\R}{\mathbb{R}}
\newcommand{\X}{\mathcal{X}}

\newcommand{\N}{\mathcal{N}}

\newcommand{\diag}{\mathrm{diag}}
\newcommand{\T}{\mathrm{T}}

\newcommand{\I}{\mathbf{I}}
\newcommand{\rk}{\mathrm{rank}}

\newcommand{\A}{\boldsymbol{A}}
\newcommand{\B}{\boldsymbol{B}}
\newcommand{\btheta}{\bar{\boldsymbol{\theta}}}

\newcommand{\Rot}{\boldsymbol{R}}

\usepackage{xcolor}
\usepackage{ifthen}
\newcommand{\markupdraft}[2]{% {#1: {color|display} command}{#2: desired color or text}
    \ifthenelse{\equal{#1}{display}}{#2}{}%                 % display only in draft version
    \ifthenelse{\equal{#1}{color}}{\color{#2}}{}%           % colored only in draft (for \new command)
}
\newcommand{\newcolored}[3][]{{\markupdraft{color}{#2}#3}%  % kept in the final print
    \ifthenelse{\equal{#1}{}}{}{\markupdraft{display}{{\color{yellow!70!black}[#1]}}}}
\newcommand{\del}[2][]{{\markupdraft{display}{{\color{orange}[removed: ``#2''[#1]]}}}} % (to be) removed
\newcommand{\calc}[2][]{{\markupdraft{display}{{\color{purple} Derived as follows: {#2}[#1]}}}} % for describe the detail in derivations.

\renewcommand{\del}[2]{}  % make removed sentences invisible
\renewcommand{\calc}[2]{}  % make removed sentences invisible
\journal{Swarm and Evolutionary Computation}

\begin{document}

\begin{frontmatter}



\title{Covariance Matrix Adaptation Evolution Strategy for Low Effective Dimensionality}


\author[label1]{Kento Uchida}
\author[label1]{Teppei Yamaguchi}
\author[label1]{Shinichi Shirakawa}

\affiliation[label1]{organization={Yokohama National University},%Department and Organization
            addressline={}, 
            city={Yokohama},
            postcode={}, 
            state={Kanagawa},
            country={Japan}}

\begin{abstract}
Despite the state-of-the-art performance of the covariance matrix adaptation evolution strategy (CMA-ES), high-dimensional black-box optimization problems are challenging tasks. Such problems often involve a property called low effective dimensionality (LED), in which the objective function is formulated with redundant dimensions relative to the intrinsic objective function and a rotation transformation of the search space. The CMA-ES suffers from LED for two reasons: the default hyperparameter setting is determined by the total number of dimensions, and the norm calculations in step-size adaptations are performed including elements on the redundant dimensions. In this paper, we incorporate countermeasures for LED into the CMA-ES and propose CMA-ES-LED. We tackle with the rotation transformation using the eigenvectors of the covariance matrix. We estimate the effectiveness of each dimension in the rotated search space using the element-wise signal-to-noise ratios of the mean vector update and the rank-$\mu$ update, both of which updates can be explained as the natural gradient ascent. Then, we adapt the hyperparameter using the estimated number of effective dimensions. In addition, we refine the cumulative step-size adaptation and the two-point step-size adaptation to measure the norms only on the effective dimensions. The experimental results show the CMA-ES-LED outperforms the CMA-ES on benchmark functions with LED.
\end{abstract}

% \includegraphics[width=0.98\textwidth]{images/graphical_abstract.pdf}

% \begin{highlights}
% \item A novel CMA-ES for problems with low effective dimensionality is proposed.
% \item Eigenvectors of covariance matrix are used to tackle the rotation of search space.
% \item The effective dimensions are estimated using element-wise SNRs of update directions.
% \item Hyperparameter setting and step-size adaptations are adaptively refined.
% \item The search performance on problems with LED was significantly improved.
% \end{highlights}


\end{frontmatter}

% ------------------------------
\section{Introduction} \label{sec:introduction}
% ------------------------------
% ------------------------------
\subsection{Background and Related Works}
% ------------------------------
The black-box optimization problem is the minimization or maximization problem in which the gradient information of the objective function is not accessible. 
These problems have appeared in several real-world applications~\cite{bboapplication}. Among the search algorithm for the black-box optimization problem with continuous search space, the covariance matrix adaptation evolution strategy (CMA-ES)~\cite{cmaes} has shown a promising search performance on several problems, containing functions that possess intractable properties such as ill-conditioned, multimodal, or non-separable landscapes. The CMA-ES employs a multivariate Gaussian distribution to generate the candidate solutions and iteratively updates the distribution parameter to generate better solutions. The update rule of the distribution parameters contains several hyperparameters. Because an improper hyperparameter setting deteriorates the search performance of CMA-ES, and because the problem-dependent hyperparameter tuning is a time-consuming task, their default settings are provided for the convenience of CMA-ES~\cite{cmaestutorial}. These default values are given by functions of the number of dimensions in the search space (i.e., the number of design variables to be optimized). 

\begin{figure*}[t]
\centering
\includegraphics[width=0.98\textwidth]{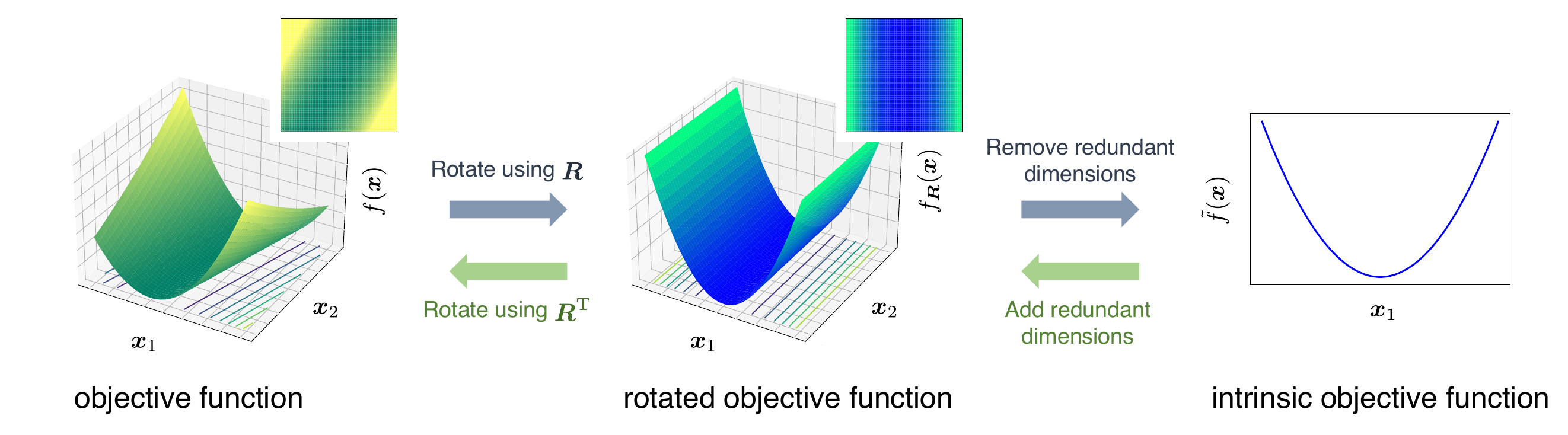}
\caption{The conceptual image of function with LED. We simply consider the case where the objective function contains a rotation matrix $\Rot \in \R^{N \times N}$ and an intrinsic objective function $\tilde{f}: \R^{\Ntrueeff} \to \R$. The objective function value at $\x$ is given by $f(\x) = \tilde{f}(\psi(\Rot \x))$, where $\psi(\y) = (\y_{1}, \cdots, \y_{\Ntrueeff})^\T \in \R^{\Ntrueeff}$. This figure shows an example with $N=2$ and $\Ntrueeff = 1$. See Section~\ref{sec:proposed} for detail.}
\label{fig:led}
\end{figure*}

Despite the state-of-the-art search performance of the CMA-ES, the high-dimensional black-box optimization problems are challenging tasks. A known intractable property of high-dimensional black-box optimization is {\it low effective dimensionality} (LED)~\cite{led}, in which the objective function value is determined by only some elements in the rotated search space and not influenced by the other elements. Problems with LED have often appeared in several real-world applications, such as the hyperparameter optimization of machine learning~\cite{mlhyperopt}, control of over-actuated systems~\cite{led-example-airplane-control}, and shape optimization~\cite{led-example-aircraft-wing-design}. Figure~\ref{fig:led} shows the conceptual image of the function with LED considered in this paper. Such objective functions contain the intrinsic objective function with a lower number of dimensions, which is not accessible. As the default hyperparameters of the CMA-ES are determined by the total number of dimensions, including redundant dimensions, LED degrades the search performance of the CMA-ES. Ideally, using the default hyperparameters value obtained by the number of the intrinsic dimensions, several update rules of the CMA-ES on the function with LED behave the same as on the intrinsic objective functions. Another weakness of the CMA-ES is the update rules in the step-size adaptations, whose performance is usually influenced by LED. The popular step-size adaptations, including the cumulative step-size adaptation (CSA)~\cite{csa1} and the two-point step-size adaptation (TPA)~\cite{tpa1}, evaluate the norm calculation by taking account of not only the effective dimensions but also the redundant dimensions. Due to these weaknesses, LED deteriorates the performance of the CMA-ES compared to the performance on the intrinsic objective function. 

Since the LED property also deteriorates other black-box optimization methods, several improvement methods have been proposed. The simplest method is to project the search space into the subspace with fewer dimensions using a random embedding. The random embedding Bayesian optimization (REMBO)~\cite{rembo} and random embedding  estimation of distribution algorithm (REMEDA)~\cite{remeda} incorporate such random embedding into Bayesian optimization and the estimation of distribution algorithm, respectively. However, random embedding contains several issues. As the number of dimensions on the intrinsic objective function is not accessible, the number of dimensions in the subspace should be chosen carefully. In addition, the random embedding may make the problem more difficult. Although several methods~\cite{ReMO:2017, REGO:2022, boundREGO:2023} using random embedding have been proposed later, those problems are not solved yet.

On the other hand, the adaptive stochastic natural gradient method for LED (ASNG-LED)~\cite{asngled} considers the effectiveness of each dimension and estimates it using the element-wise signal-to-noise ratio (SNR) of the update direction of the distribution parameter. ASNG-LED incorporates this mechanism into an adaptation method of the learning rate~\cite{asng} for the stochastic natural gradient. ASNG-LED successfully improves the search performance of ASNG on binary optimization problems with LED. Although the stochastic natural gradient recovers some of the update rules of the CMA-ES~\cite{bidirectional}, this approach cannot be incorporated directly because the rank-one update and the step-size adaptation are not recovered. Moreover, in continuous search space, because the projection between the search spaces of the objective function and the intrinsic objective function often involves a rotation transformation, the estimation of the rotation transformation is additionally required.

% ------------------------------
\subsection{Our Contributions}
% ------------------------------

In this paper, we propose an estimation method of the effectiveness of each dimension and the rotation transformation, which reconstructs the landscape of the intrinsic objective function. Firstly, we estimate the rotation transformation using the eigenvectors of the covariance matrix in the CMA-ES. Then, we calculate the rotated update direction of the mean vector and the covariance matrix and estimate their element-wise SNRs. To achieve the effectiveness of each dimension from the element-wise SNRs, we introduce a monotonically increasing function with two tunable parameters. These parameters are adaptively updated based on the number of dimensions, the sample size, and the maximum element of element-wise SNRs, where any problem-dependent tuning by the user is unnecessary.

Based on the estimated effectiveness of each dimension and the rotation transformation, we incorporate two countermeasures for LED into the CMA-ES and propose the CMA-ES-LED. The first is the hyperparameter adaptation using the default hyperparameter settings of the CMA-ES, in which we compute the hyperparameter values using the estimated number of effective dimensions instead of the total number of dimensions. The second is the refinement of the norm calculation in well-known step-size adaptations, the CSA and the TPA. We compute the norm of the evolution path and a random noise using the effectiveness of dimensions as the weight. We note that, with ideal estimation of effective dimensions, the dynamics of the CMA-ES-LED on the objective function with LED are identical to the dynamics of the CMA-ES on the intrinsic objective function. 

The experimental results show that CMA-ES-LED performs significantly better than the original CMA-ES on the benchmark functions with LED. At the same time, the CMA-ES-LED is competitive with the CMA-ES on functions without LED. Additionally, we incorporate the IPOP restart strategy~\cite{ipopcmaes} into CMA-ES-LED to investigate the search performance on multimodal functions, which demonstrates the improvements of CMA-ES-LED over the CMA-ES in the cases of LED.

This study is an extension of~\cite{cmaesled}, in which the estimation mechanism of the effectiveness of each dimension and the same countermeasures for LED are applied to the sep-CMA-ES~\cite{sepcmaes}. The sep-CMA-ES is a variant of CMA-ES and restricts the covariance matrix to a diagonal matrix. We note that, differently from CMA-ES-LED, the methods in~\cite{cmaesled} cannot handle the rotation transformation of the search space. The estimation of the rotation transformation is one of the novelties of this work, which allows CMA-ES-LED to inherit the invariance properties of the CMA-ES.

% ------------------------------
\subsection{Organization of This Paper}
% ------------------------------

This paper is organized as follows. Section~\ref{sec:cmaes} describes the CMA-ES as our baseline algorithm. In Section~\ref{sec:proposed}, we introduce the estimation process and the countermeasures for LED applied to the CMA-ES-LED. Section~\ref{sec:experiment} shows the result of the numerical simulations to evaluate the search performance of the CMA-ES-LED. Finally, we conclude this paper in Section~\ref{sec:conclusion}.

% ------------------------------
\section{Covariance Matrix Adaptation Evolution Strategy} \label{sec:cmaes}
\subsection{Algorithm of CMA-ES}
% ------------------------------
The covariance matrix adaptation evolution strategy (CMA-ES)~\cite{cmaes} is a black-box optimization method for continuous variables. Let us consider the minimization of $N$-dimensional unconstrained objective function $f: \R^N \to \R$.
The CMA-ES employs a multivariate Gaussian distribution as the search distribution and updates its parameters to generate superior solutions.
The Gaussian distribution $\N(\mv^{(t)}, (\sigma^{(t)})^2 \cov^{(t)})$ is parametrized by the mean vector $\mv^{(t)} \in \R^N$, covariance matrix $\cov^{(t)} \in \R^{N \times N}$, and step-size $\sigma^{(t)} \in \R_{>0}$. 

The single update of the CMA-ES is as follows; 
% First, the CMA-ES generates $\lambda$ candidate solutions $\x_k = \mv^{(t)} + \sigma^{(t)} \y_k$ for $k = 1, \cdots, \lambda$ using samples $\y_k$ generated from $\N(\boldsymbol{0}, \cov^{(t)})$.
First, the CMA-ES generates $\lambda$ candidate solutions $\x_1, \cdots, \x_\lambda$ as
\begin{align}
    \y_k &= \sqrt{\cov^{(t)}} \z_k \qquad\quad \text{with} \quad \z_k \sim \N(\boldsymbol{0}, \I)\\
    \x_k &= \mv^{(t)} + \sigma^{(t)} \y_k
\end{align}
for $k = 1, \cdots, \lambda$.
The candidate solutions are then evaluated on the objective function and sorted by their ranking. The index of $i$-th best candidate solution is written as $i:\lambda$, i.e., it satisfies $f(\x_{1:\lambda}) \leq \cdots \leq f(\x_{\lambda:\lambda})$. Introducing the decreasing positive weights $w_1 > \cdots > w_\mu > 0$, the weighted average of the best $\mu$ samples $\y_{1:\lambda}, \cdots \y_{\mu:\lambda}$ is calculated as
\begin{align}
        \langle \y \rangle_w^{(t+1)}  = \sum_{i=1}^{\mu} w_i \y_{i:\lambda} \enspace.
\end{align}
The weights are given by $w_i = {w_i'} / \left( {\sum^\lambda_{j=1} w_j'} \right)$, where $w_i'$ is set as
\begin{align}
        w_i' = \max\left( \ln \frac{\lambda + 1}{2} - \ln i, 0 \right) \enspace.
\end{align}
Then, the update direction $\Delta \mv^{(t+1)}$ of the mean vector reads
\begin{align}
    \Delta \mv^{(t+1)} = \sigma^{(t)} \langle \y \rangle_w^{(t+1)} \enspace. 
    \label{eq:mean-update-direct}
\end{align}
The update rule of the mean vector is given by
\begin{align}
        \mv^{(t+1)} = \mv^{(t)} + c_m \Delta \mv^{(t+1)} \enspace, 
        \label{eq:mean-update} 
\end{align}
where $c_m > 0$ is the learning rate, which is usually set as $c_m = 1$.

The update rule of the covariance matrix consists of two updates: the rank-$\mu$ update and the rank-one update. In the rank-$\mu$ update, the covariance matrix is updated to the weighted sample covariance of the best $\mu$ candidate solutions. The update direction of the rank-$\mu$ update is given by 
\begin{align}
        \Delta_{\mu} \cov^{(t+1)} = \sum_{i=1}^{\mu} w_i \left( \y_{i:\lambda} \y_{i:\lambda}^\T - \cov^{(t)} \right) \enspace. 
        \label{eq:rank-mu-update}
\end{align}
The rank-one update, on the other hand, elongates the covariance matrix along the mean vector update direction. The CMA-ES introduces the evolution path $\p_c^{(t)} \in \R^N$ to accumulate the update direction of the mean vector (divided by $\sigma^{(t)}$) with the accumulation factor $c_c > 0$ as
\begin{align}
        \qquad \p_c^{(t+1)} = (1 - c_c) \p_c^{(t)} 
        + h_\sigma^{(t)} \sqrt{c_c (2-c_c) \mueff} \frac{\Delta \mv^{(t+1)}}{\sigma^{(t)}} \enspace,
        \label{eq:rank-one-path-update}
\end{align}
where the initial value is given by $\p_{c}^{(0)} = \mathbf{0}$.
The parameters $\mueff=(\sum_{i=1}^{\mu}w_i^2)^{-1}$ and $h_\sigma^{(t)}$ are the variance effective selection mass and Heaviside function, respectively. 
The Heaviside function takes $h_\sigma^{(t)} = 1$ (usually) or $h_\sigma^{(t)} = 0$ (unusually). 
The setting of the Heaviside function depends on the update rule of the step-size. In general, it takes $h_\sigma^{(t)} = 0$ when $\sigma^{(t)}$ increases dramatically, which stalls the the update of $\p_c^{(t)}$.
The update direction of the rank-one update reads
\begin{align}
        \Delta_1 \cov^{(t+1)} &=  \p_c^{(t+1)} \left( \p_c^{(t+1)} \right)^\T - \cov^{(t)} \enspace.
        \label{eq:rank-one-update}
\end{align}
Totally, with the learning rates $c_\mu$ and $c_1$ for the rank-$\mu$ update and the rank-one update, the covariance matrix is updated as
\begin{multline}
        \cov^{(t+1)} = (1 + (1 - h_\sigma^{(t)}) c_1 c_c(2-c_c)) \cov^{(t)} \\
        + c_\mu \Delta_\mu \cov^{(t+1)} + c_1 \Delta_1 \cov^{(t+1)} \enspace. 
        \label{eq:cov-update}
\end{multline}

Because the update of step-size, called as {\it the step-size adaptation}, is critical to the search performance, several update rules have been proposed. We introduce two well-known step-size adaptations, the CSA~\cite{csa1} and the TPA~\cite{tpa1}, as follows:

% ------------------------------
\subsubsection{Cumulative Step-size Adaptation (CSA)}
% ------------------------------
The update rule of the CSA is based on the dynamics of the mean vector. When the mean vector moves toward a certain direction, the increase in the step-size improves the search efficiency. When the mean vector stays around the same position, on the other hand, a decrease in the step-size improves the local search ability.
Based on this reason, the CSA employs another evolution path $\p_{\sigma}^{(t)} \in \R^N$, which is initialized as $\p_{\sigma}^{(0)} = \mathbf{0}$ and accumulates the update direction $\Delta \mv^{(t+1)}$ as
\begin{align}
        \p_{\sigma}^{(t+1)} &= (1-c_\sigma)\p_{\sigma}^{(t)} + \sqrt{ c_\sigma(2-c_\sigma)\mueff} \langle \z \rangle_w^{(t+1)} \enspace, 
\end{align}
where $c_\sigma > 0$ is the accumulation factor and
\begin{align}
        \langle \z \rangle_w^{(t+1)} = \left( \cov^{(t)} \right)^{- \frac{1}{2}} \frac{\Delta \mv^{(t+1)}}{\sigma^{(t)}} \enspace. 
\end{align}
The CSA updates the step-size based on the norm of evolution path $\| \p_{\sigma}^{(t+1)} \|$ as
\begin{align}
        \sigma^{(t+1)} = \sigma^{(t)} \exp\left(\frac{c_\sigma}{d_\sigma}\left(\frac{\|\p_\sigma^{(t+1)}\|}{\E[\|\N(\0, \I)\|]}-1\right)\right) \enspace, 
        \label{eq:csa}
\end{align}
where $d_\sigma > 0$ is the damping factor. The Heaviside function for the CSA becomes one, i.e., $h^{(t)} = 1$, when
\begin{align}
        \frac{ \|\p_\sigma^{(t+1)}\| }{\sqrt{ 1 - (1 - c_\sigma )^{2 (t + 1)} }} < \left( 1.4 + \frac{2}{N + 1} \right) \E \left[ \| \mathcal{N}(\0, \I) \| \right] \enspace. 
        \label{eq:csa-heaviside}
\end{align}
For the expectation of the norm $\| \N(\0, \I) \|$, we use a well-known approximated value as
\begin{align}
        \E[\|\N(\0, \I)\|] \approx \sqrt{N} \left(1 - \frac{1}{4N} + \frac{1}{21 N^2} \right) \enspace. 
\end{align}

% ------------------------------
\subsubsection{Two-Point Step-Size Adaptation (TPA)}
% ------------------------------
The update procedure of the TPA works as the line search along the update direction of the mean vector $\Delta \mv^{(t)}$.
In the TPA, two additional candidate solutions $\x_{+}$ and $\x_{-}$ are generated symmetrically along $\Delta \mv^{(t)}$ as
\begin{align}
        \x_{\pm} = \mv^{(t)} \pm \frac{\sigma^{(t)} \|\N(\0, \I)\| \cdot \Delta \mv^{(t)}}{ \sqrt{ (\Delta \mv^{(t)})^\T (\cov^{(t)})^{-1} \Delta \mv^{(t)} }} \enspace. 
        \label{eq:tpa-sample}
\end{align}
Then, two candidate solutions are replaced with $\x_{+}$ and $\x_{-}$ not to change the sample size.
When $\x_{+}$ is superior to $\x_{-}$ on $f$, increasing the step-size is reasonable because better solutions may be found beyond the mean vector. Otherwise, the decrease in the step-size promotes local search around the mean vector. Based on this principle, the TPA accumulates the difference between the rankings of $\x_{+}$ and $\x_{-}$ as
\begin{align}
        s^{(t+1)} = (1-c_\sigma) s^{(t)} + c_\sigma \frac{\rk(\x_{-}) - \rk(\x_{+})}{\lambda - 1} \enspace,
        \label{eq:tpa-accumulation}
\end{align}
where $\rk(\x)$ returns the ranking of $\x$ among $\lambda$ samples.
Then, the TPA updates the step-size as
\begin{align}
        \sigma^{(t+1)} = \sigma^{(t)} \exp\left( \frac{ s^{(t+1)} }{ d_\sigma } \right) \enspace. 
        \label{eq:tpa}
\end{align}
The Heaviside function is set as $h_\sigma^{(t)} = \mathbb{I}\{ s^{(t+1)} < 0.5 \}$, which is introduced in~\cite{vkdcmaes}.

% ------------------------------
\subsection{Default Hyperparameter Settings} \label{sec:hyperparameter}
% ------------------------------
The setting of hyperparameters influences the search performance of the CMA-ES. Tuning the hyperparameter is usually tedious, although it may improve the search performance.
To reduce the tuning cost, the default hyperparameter setting is provided~\cite{cmaestutorial}. These default values are given by functions of the number of dimensions $N$ of the search space and the sample size $\lambda$.\footnote{As the default setting of the sample size $\lambda = 4 + \lfloor 3 \ln N \rfloor$ is also a function of $N$, all default values are determined by $N$.}
The default settings of the hyperparameters $c_c$, $c_1$, and $c_\mu$ for the covariance matrix update are set as
\begin{align}
\begin{aligned}
    c_c &= \frac{4 + \mueff / N}{N + 4 + 2\mueff / N} \enspace  \\
    c_1 &= \frac{2}{(N+1.3)^2 + \mueff} \enspace  \\
    c_\mu &= \min\left(1-c_1, \frac{2(\mueff-2+1/\mueff)}{(N+2)^2+\mueff}\right) \enspace. \\
\end{aligned} \label{eq:hyperparameter}
\end{align}
The hyperparameters $c_\sigma$ and $d_\sigma$ for the CSA and the TPA have different default settings. For the CSA, $c_\sigma$ and $d_\sigma$ are set as
\begin{align}
\begin{aligned}
c_\sigma &= \frac{\mueff + 2}{N + \mueff + 5} \\
d_\sigma &= 1 + c_\sigma + 2\max\left( 0, \sqrt{\frac{\mueff - 1}{N + 1}} - 1 \right) \enspace. \\
\end{aligned} \label{eq:hyperparameter-csa} 
\end{align}
In contrast, the default setting for the TPA reads
\begin{align}
c_\sigma = 0.3 \qquad \text{and} \qquad
d_\sigma = \sqrt{N} \enspace.\label{eq:hyperparameter-tpa} 
\end{align}

% ------------------------------
\subsection{Invariance Properties of CMA-ES} \label{sec:invaraince}
% ------------------------------
Invariance properties of the search algorithm ensure that its behaviors are identical when the corresponding transformation is applied to the search space or objective function. Invariance properties make the search performance of an algorithm robust. The CMA-ES possesses several invariance properties as follows.
\begin{itemize}
\item Invariance to any invertible linear transformation of the search space. Precisely, for any invertible matrix $\A \in \R^{N \times N}$ and any vector $\boldsymbol{b} \in \R^N$, the dynamics of $(\mv^{(t)}, \cov^{(t)}, \sigma^{(t)})$ on $f(\x)$ is identical to $(\A \mv^{(t)} + \boldsymbol{b}, \A \cov^{(t)} \A^\T , \sigma^{(t)})$ on $f_\mathrm{linear}(\x) : \x \mapsto f( \A \x + \boldsymbol{b} )$ if the corresponding initial state is given.
Particularly, setting $\A$ be an arbitrary permutation matrix and $\boldsymbol{b} = \mathbf{0}$ holds invariance to any permutation of the order of the design variables.
\item Invariance to any order-preserving transformation of the objective function value. For any strictly increasing $g: \R \to \R$, the behaviors of the CMA-ES on $f$ and $f_\mathrm{order}: \x \mapsto g(f( \x ))$ are identical.
\end{itemize}
Compared to the previous work~\cite{cmaesled}, our proposed method aims to inherit these invariance properties of the CMA-ES, including the invariance to any rotation transformation.

% ------------------------------
\subsection{Relation to Stochastic Natural Gradient Method}
% ------------------------------
The mean vector update and the rank-$\mu$ update in the CMA-ES closely relate to the stochastic natural gradient method (SNG)~\cite{igo}.
The SNG employs a family of probability distributions $\{ P_{\boldsymbol{\theta}} \}$ parameterized by $\boldsymbol{\theta} \in \mathrm{\Theta}$ on the search space $\X$ and transforms the original problem to the maximization of the expectation of the utility function\footnote{The utility function assigns a higher value to a better solution, which is a nonlinear and non-increasing transformation of the objective function $f$.} $u: \R^N \to \R$ as
\begin{align}
    \max_{\boldsymbol{\theta} \in \Theta} \enspace J(\boldsymbol{\theta}) \quad \text{where} \quad J(\boldsymbol{\theta}) = \int_{\x \in \X} u(\x) p_{\boldsymbol{\theta}} (\x) \mathrm{d} \x \enspace, 
\end{align}
where $p_{\boldsymbol{\theta}}$ is the probability density function of $P_{\boldsymbol{\theta}}$.
The SNG updates the distribution parameter along the natural gradient direction of $J(\boldsymbol{\theta})$. The natural gradient is the steepest direction w.r.t. the Fisher metric~\cite{naturalgradient} and given by $\tilde{\nabla}_{\boldsymbol{\theta}} J(\boldsymbol{\theta}) = F^{-1}(\boldsymbol{\theta}) \nabla_{\boldsymbol{\theta}} J(\boldsymbol{\theta})$, where $F^{-1}(\boldsymbol{\theta})$ indicates the inverse of the Fisher information matrix.
Applying the log-likelihood trick ${\nabla}_{\boldsymbol{\theta}} p_{\boldsymbol{\theta}}(\x) = ({\nabla}_{\boldsymbol{\theta}} \ln p_{\boldsymbol{\theta}}(\x)) p_{\boldsymbol{\theta}}(\x)$, the natural gradient is approximated by Monte Carlo estimation using $\lambda$ samples generated from $P_{\boldsymbol{\theta}}$ as
\begin{align}
    \tilde{\nabla}_{\boldsymbol{\theta}} J(\boldsymbol{\theta}) &\approx \frac{1}{\lambda} \sum^\lambda_{i=1} u(\x_i) \tilde{\nabla}_{\boldsymbol{\theta}} \ln p_{\boldsymbol{\theta}}(\x_i) \enspace,
\end{align}
where $\tilde{\nabla}_{\boldsymbol{\theta}} \ln p_{\boldsymbol{\theta}}(\x_i)$ is the natural gradient of log-likelihood.
Introducing the learning rate $\eta > 0$, the update rule of the SNG is derived as
\begin{align}
    \boldsymbol{\theta}^{(t+1)} = \boldsymbol{\theta}^{(t)} + \frac{\eta}{\lambda} \sum^\lambda_{i=1} u(\x_i) \left. \tilde{\nabla}_{\boldsymbol{\theta}} \ln p_{\boldsymbol{\theta}}(\x_i) \right|_{\boldsymbol{\theta} = \boldsymbol{\theta}^{(t)}} \enspace.
\end{align}

When applying a multivariate Gaussian distribution $\N(\mv, \cov)$ parametrized by the mean vector $\mv$ and the covariance matrix $\cov$ and setting $u(\x_{i:\lambda}) = w_i$, the estimated natural gradients w.r.t. $\mv$ and $\cov$ are given by $\Delta \mv$ in \eqref{eq:mean-update-direct} and $\Delta_\mu \cov$ in \eqref{eq:rank-mu-update}, respectively.
This relationship helps us to understand the design principle of our proposed method.

% ------------------------------
\section{Problem with Low Effective Dimensionality} \label{sec:problem}
% ------------------------------

We define the problem with low effective dimensionality~(LED). We consider the intrinsic objective function $\tilde{f}: \R^{\Ntrueeff} \to \R$, where $\Ntrueeff < N$ is the number of effective dimensions on $\tilde{f}$. 
Here, we call $i$-th dimension an effective dimension on $f$ when there exist $\delta \in \R$ and $\x \in \X$ such that replacing the $i$-th element $\x_i$ of input $\x$ with $\x_i + \delta$ changes the evaluation value on $f$.
We also consider a rotation matrix $\Rot \in \R^{N \times N}$ that is not accessible. Then, the target objective function $f: \R^N \to \R$ is constructed as
\begin{align}
        &f(\x) = \tilde{f}( \psi( \Rot \x )) \\
        &\text{where} \quad \psi(\y) = \left( \y_{1}, \cdots, \y_{\Ntrueeff} \right)^\T \enspace.
\end{align}
Figure~\ref{fig:led} depicts the conceptual image of our problem setting. We note that the target objective function has $N$ effective dimensions (except for some trivial cases such as $\Rot = \I$) while the intrinsic objective function has $\Ntrueeff$ ones.

% ------------------------------
\section{CMA-ES for Low Effective Dimensionality} \label{sec:proposed}
% ------------------------------

To demonstrate the desired dynamics of the distribution parameters ${\mv}$, ${\cov}$, and ${\sigma}$ on $f$, we compare them with the dynamics of the distribution parameters $\tilde{\mv}$, $\tilde{\cov}$, and $\tilde{\sigma}$ on $\tilde{f}$. 
We consider the case that the initial distribution parameters on $\tilde{f}$ are set as
\begin{align}
    \tilde{\mv}^{(0)}_{i} &= ( \Rot \mv^{(0)} )_{i} \enspace, \quad 
    \tilde{\cov}^{(0)}_{i,j} = ( \Rot \cov^{(0)} \Rot^\T )_{i,j} \enspace, \quad 
    \tilde{\sigma}^{(0)} = \sigma^{(0)} 
\end{align}
for all $i,j \in \{1, \cdots, \Ntrueeff\}$, where $\A_{i,j}$ denotes the $(i,j)$ element of a matrix $\A$.
We assume the same hyperparameters and random noises $\{ \z_k \}_{k=1}^\lambda$ from the standard Gaussian distribution $\N(\0, \I)$ are given. Then, if the dynamics of the step-size $\sigma$ is the same, it satisfies
\begin{align}
        \tilde{\mv}^{(t)}_i &= ( \Rot \mv^{(t)} )_{i} \quad \text{and} \quad
        \tilde{\cov}^{(t)}_{i,j} = ( \Rot \cov^{(t)} \Rot^\T )_{i,j}
\end{align}
for all $t > 0$. Moreover, the dynamics of the best evaluation value on $f$ and $\tilde{f}$ are also the same. This means that the hyperparameter settings and the updates of the step-size $\sigma$ on $f$ and $\tilde{f}$ should be the same to realize the same behavior on both  $f$ and $\tilde{f}$. 
However, because the default hyperparameters in \eqref{eq:hyperparameter}, \eqref{eq:hyperparameter-csa}, and \eqref{eq:hyperparameter-tpa} are functions of the number of dimensions, they are different on $f$ and $\tilde{f}$. In addition,
the CSA and the TPA work differently because the norms are measured taking account of the redundant dimensions. Due to these factors, performance deterioration of the CMA-ES occurs when $N \gg \Ntrueeff$.

We introduce a rotation matrix $\tilde{\Rot} \in \R^{N \times N}$ to explain the design principle of the proposed method. We consider the rotated objective function by $\tilde{\Rot}$ as
\begin{align}
        f_{\tilde{\Rot}}(\x) = f(\tilde{\Rot}^\T \x) = \tilde{f}( \psi( \Rot \tilde{\Rot}^\T \x )) \enspace. \label{eq:rot-f}
\end{align}
% We note $f_{\Rot}(\x) = \tilde{f}( \psi( \x ))$, which is shown in the middle of Figure~\ref{fig:led}.
% Due to the invariance to any invertible affine transformation of the search space, the dynamics of the distribution parameters on $f$ are corresponding to those on $f_{\tilde{\Rot}}$.
% Then, we consider the case where $\tilde{\Rot}$ makes $f_{\tilde{\Rot}}$ consist of $\Ntrueeff$ effective dimensions. 
% Given $\Ntrueeff$ effective dimensions on $f_{\tilde{\Rot}}$ with such $\tilde{\Rot}$, we can update the step-size $\sigma$ on $f$ (and $f_{\tilde{\Rot}}$) as the same as $\tilde{\sigma}$ on $\tilde{f}$ and can set the hyperparameters as on $\tilde{f}$, which leads the same search performance on $f$ and $\tilde{f}$.
If $\tilde{\Rot} = \Rot$, the rotated objective function is given by $f_{\Rot}(\x) = \tilde{f}( \psi( \x ))$ and clearly contains $\Ntrueeff$ effective dimensions, i.e., only $\x_1, \cdots,  \x_{\Ntrueeff}$ affect the evaluation value on $f_{\Rot}$. We note that $\Rot$ is not a unique rotation matrix to make $f_{\tilde{\Rot}}$ consist of $\Ntrueeff$ effective dimensions, as discussed in Section~\ref{sec:est-eff-dim}.

The aim of this paper is to propose countermeasures to tackle such performance deterioration on $f$.
In this section, we firstly introduce a reasonable choice for the rotation matrix $\tilde{\Rot}$. We then estimate the element-wise signal-to-noise ratio of update direction on rotated search space by $\tilde{\Rot}$.
Then, we incorporate two following countermeasures into CMA-ES and propose a variant of CMA-ES, termed CMA-ES-LED:
\begin{itemize}
\item A hyperparameter adaptation mechanism based on the estimated number of effective dimensions.
\item Refinements of the update rules of the CSA and the TPA to measure the norms only on the effective dimensions.
\end{itemize}

% ------------------------------
\subsection{Estimation of Effectiveness of Dimensions} \label{sec:est-eff-dim}
% ------------------------------
As introduced in \cite{asngled, cmaesled}, we introduce an $N$-dimensional vector $\vv^{(t)} \in [0, 1]^{N}$ that represents the estimated effectiveness of each dimension on $f_{\tilde{\Rot}}$. Our estimation aims to make the elements of $\vv^{(t)}$ corresponding to the effective dimensions closer to one and to make the other elements closer to zero. 

\paragraph{Estimation of Rotation Matrix}
Here, we consider the condition for $\tilde{\Rot}$ to make the rotated objective function $f_{\tilde{\Rot}}$ involve $\Ntrueeff$ effective dimensions and $N-\Ntrueeff$ redundant dimensions. The condition for $\tilde{\Rot}$ is as follows: there are an arbitrary permutation matrix $\boldsymbol{P} \in \R^{N \times N}$ and arbitrary rotation matrices $\boldsymbol{D}_\text{eff} \in \R^{\Ntrueeff \times \Ntrueeff}$ and $\boldsymbol{D}_\text{red} \in \R^{(N - \Ntrueeff) \times (N - \Ntrueeff)}$ satisfying
\begin{align}
        \Rot \tilde{\Rot}^\T = \left( \begin{array}{cc}
                \boldsymbol{D}_\text{eff} & \boldsymbol{O} \\
                \boldsymbol{O} & \boldsymbol{D}_\text{red}
            \end{array} \right) \boldsymbol{P} \enspace. \label{eq:cond}
\end{align}
Inserting this into \eqref{eq:rot-f} proofs the statement.

\begin{figure}[t]
\centering
\includegraphics[width=0.45\textwidth]{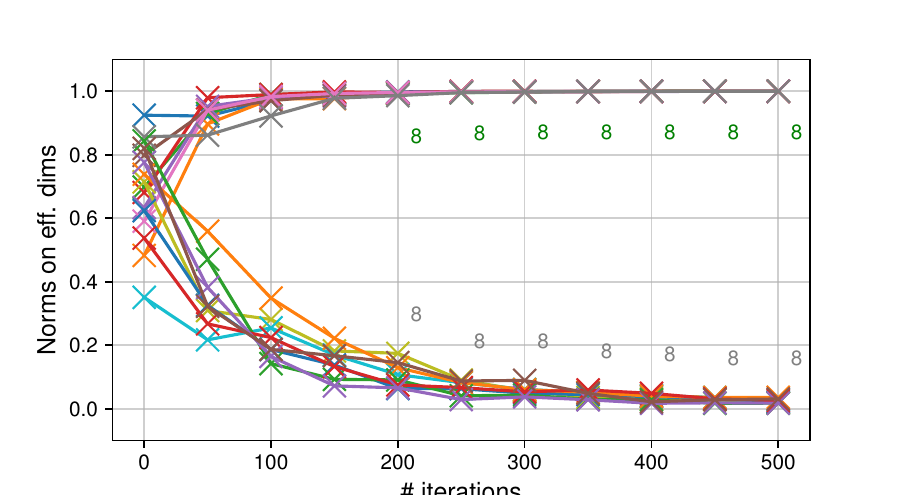}
\caption{The transitions of the norms of rotated eigenvectors of the covariance matrix on the effective dimensions $\| \bar{\boldsymbol{b}}_i \|_\mathrm{eff}$. We also plot the number of lines above and below 0.5. This is a typical result of CMA-ES with the CSA on the Sphere function. We set $N = 16$ and $\Ntrueeff=8$. The rotation matrix $\Rot$ was randomly given.}
\label{fig:eigvec_trans}
\end{figure}

As we cannot access $\Rot$ in practice, we consider a rotation matrix which does not require $\Rot$ and approximately satisfies the condition \eqref{eq:cond}. We choose the eigenvectors $\B^{(t)}$ of the covariance matrix $\cov^{(t)}$, which is obtained by the eigendecomposition $\cov^{(t)} = \B^{(t)} \boldsymbol{\Lambda}^{(t)} (\B^{(t)})^\T$, where $\boldsymbol{\Lambda}^{(t)}$ is the diagonal matrix whose diagonal elements are the eigenvalues of $\cov^{(t)}$. To demonstrate whether $\B^{(t)}$ approximately satisfies the condition \eqref{eq:cond}, we compute the norms of column vectors in $\Rot (\B^{(t)})^\mathrm{T}$ on effective dimensions in optimization. The norm of the $i$-th column vector $\bar{\boldsymbol{b}}_i \in \R^N$ in the matrix $\Rot (\B^{(t)})^\mathrm{T}$ computed on effective dimensions is 
\begin{align}
    \| \bar{\boldsymbol{b}}_i \|_\mathrm{eff} = \sqrt{ \sum^{\Ntrueeff}_{j=1} \boldsymbol{\bar{b}}^2_{i, j} } \enspace.
\end{align}
If $\B^{(t)}$ approximately satisfies \eqref{eq:cond}, the norms of $\Ntrueeff$ column vectors become close to one, and the norms of the rest $N - \Ntrueeff$ column vectors become close to zero. Figure~\ref{fig:eigvec_trans} shows the transitions of these norms on Sphere function with $N=16$ and $\Ntrueeff=8$ (see the definition of benchmark functions with LED in Table~\ref{table:benchmark-functions}). As we expected, only $\Ntrueeff$ norms increased to one, and the other norms decreased to zero.

\paragraph{Estimation of element-wise SNR}
Similarly to~\cite{asngled, cmaesled}, we update $\vv^{(t)}$ using the element-wise signal-to-noise ratios (SNRs) of the estimated natural gradients $\Delta \mv$ and $\Delta_\mu \cov$. With the rotation matrix $\tilde{\Rot}$, the target element-wise SNRs are defined as
\begin{align}
        \frac{ \left( \E \left[ ( \tilde{\Rot}^\T \Delta \mv^{(t+1)})_i  \right] \right)^2 }{ \Var \left[ ( \tilde{\Rot}^\T \Delta \mv^{(t+1)})_i \right] } 
        \,, \,
        \frac{ \left( \E\left[ ( \tilde{\Rot}^\T \Delta_\mu \cov^{(t+1)} \tilde{\Rot})_{i,i} \right] \right)^2 }{ \Var \left[ ( \tilde{\Rot}^\T \Delta_\mu \cov^{(t+1)} \tilde{\Rot} )_{i,i} \right] } \enspace. \label{eq:element-snr}
\end{align}
These element-wise SNRs are corresponding to the coordinate-wise SNRs of $\Delta \mv^{(t+1)}$ and $\Delta_\mu \cov^{(t+1)}$ on $f_{\tilde{\Rot}}$ that involve $\Ntrueeff$ effective dimensions with $\tilde{\Rot}$ satisfying \eqref{eq:cond}. The element-wise SNRs are zero on the redundant dimensions because the elements of the natural gradient are zero. In contrast, the element-wise SNR tend to be large on the effective dimensions. Therefore, we estimate the effective dimensions using the element-wise SNRs.

In practice, these element-wise SNRs cannot be derived analytically. To estimate them, we introduce the following accumulations using $\B^{(t)}$ instead of $\tilde{\Rot}$ as
\begin{align}
        \accs^{(t+1)}_{\mv, i} & = (1 - \beta) \accs_{\mv, i}^{(t)} + \sqrt{\beta(2-\beta)} \Delta \bar{\mv}^{(t+1)}_i \label{eq:accs-mv-update} \\
        \accgamma^{(t+1)}_{\mv, i} & = (1 - \beta)^2 \accgamma_{\mv, i}^{(t)} + \beta(2-\beta) \left( \Delta \bar{\mv}^{(t+1)}_i \right)^2 \label{eq:accgamma-mv-update} \\
        \accs^{(t+1)}_{\cov, i} & = (1 - \beta) \accs_{\cov, i}^{(t)} + \sqrt{\beta(2-\beta)}  \Delta \bar{\cov}^{(t+1)}_{i} \label{eq:accs-cov-update} \\
        \accgamma^{(t+1)}_{\cov, i} & = (1 - \beta)^2 \accgamma_{\cov, i}^{(t)} + \beta(2-\beta) \left( \Delta \bar{\cov}^{(t+1)}_{i} \right)^2 \enspace, \label{eq:accgamma-cov-update}
\end{align}
where $\beta \in (0,1]$ is the smoothing factor and
\begin{align}
        \Delta \bar{\mv}^{(t+1)} &= ( \B^{(t)} )^\T \Delta \mv^{(t+1)} \label{eq:bar-mv} \\
        \Delta \bar{\cov}^{(t+1)} &= \diag^\ast( (\B^{(t)})^\T \Delta_\mu \cov^{(t+1)} \B^{(t)} ) \enspace.  \label{eq:bar-cov}
\end{align}
The operation $\diag^\ast$ returns the diagonal elements of the inputted matrix.
We note that introducing $\B^{(t)}$ maintains the rotation invariance of the proposed method.

\begin{figure*}[t]
\centering
\includegraphics[width=0.6\textwidth]{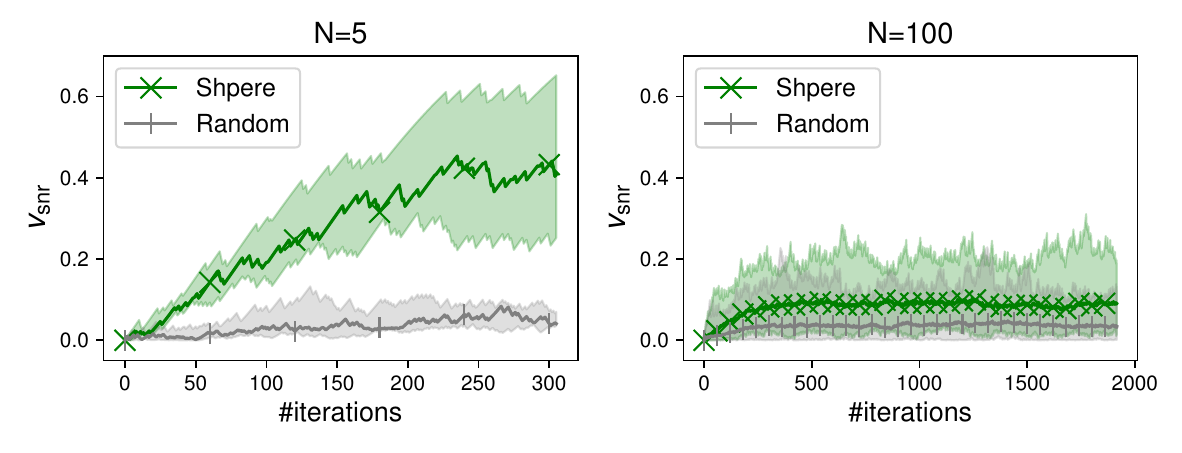}
\caption{The transitions of the elements of $\vsnr$ on the sphere function (green) and the random function (gray). The solid lines and shaded areas show the median and ranges between the minimum and maximum, respectively. Note that these lines are obtained by a single trial of the CMA-ES with the CSA.}
\label{fig:snr}
\end{figure*}
\begin{figure*}[t]
\centering
\includegraphics[width=0.6\textwidth]{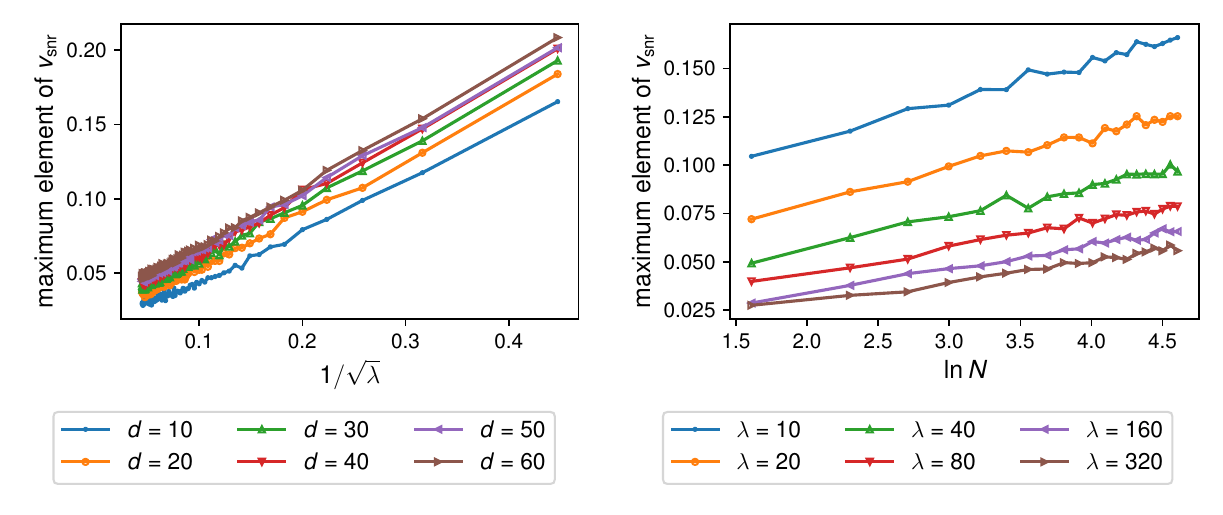}
        \caption{
        The maximum element of $\vsnr$ from 1,000-th iteration to 2,000-th iteration in the CMA-ES on the random function.
        The average values over ten runs are displayed.
        }
\label{fig:maximum-snr}
\end{figure*}

Here, we consider the case where the learning rates are so small that the distribution parameters stay around the same point for $\tau$ iterations. Then, we can approximately transform the expected values of the accumulations $\accs^{(t+1)}_{\btheta, i}$ and $\accgamma^{(t+1)}_{\btheta, i}$, where $\btheta \in \{\bar{\mv}, \bar{\cov}\}$, as
\begin{align}
        \E\left[ \accs^{(t+1)}_{\btheta, i} \right] &\approx \sqrt{\beta (2 - \beta)} \E\left[ \Delta \btheta_i^{(t+1)} \right] \sum^{\tau}_{k=0} (1 - \beta)^k \\
        &\overset{\tau \to \infty}{\longrightarrow} \sqrt{ \frac{2 - \beta}{\beta}} \E\left[ \Delta \btheta_i^{(t+1)} \right] \label{eq:accs-expectation} \\
        \E\left[ \accgamma^{(t+1)}_{\btheta, i} \right] &\approx \beta (2 - \beta) \E\left[  \left( \Delta \btheta_i^{(t+1)} \right)^2 \right] \sum^{\tau}_{k=0} (1 - \beta)^{2k} \\
        &\overset{\tau \to \infty}{\longrightarrow} \E\left[  \left( \Delta \btheta_i^{(t+1)} \right)^2 \right] \enspace. \label{eq:accgamma-expectation}
\end{align}
Similarly, we can transform the variance of $\accs^{(t+1)}_{\btheta, i}$ as
\begin{align}
    \Var\left[ \accs^{(t+1)}_{\btheta, i} \right] &\approx \beta (2 - \beta) \Var\left[ \Delta \btheta_i^{(t+1)} \right]  \sum^{\tau}_{k=0} (1 - \beta)^{2k} \\
    &\overset{\tau \to \infty}{\longrightarrow} \Var\left[ \Delta \btheta_i^{(t+1)} \right] \enspace.\label{eq:accs-variance}
\end{align}
We consider to estimate the element-wise SNRs~\eqref{eq:element-snr} using the expectations of $(\accs^{(t+1)}_{\btheta, i})^2$ and $\accgamma^{(t+1)}_{\btheta, i}$.
Using approximations \eqref{eq:accs-expectation}, \eqref{eq:accgamma-expectation} and \eqref{eq:accs-variance}, we obtain 
\begin{align}
        &\E\left[ \left( \accs^{(t+1)}_{\btheta, i} \right)^2 \right] - \E\left[ \accgamma^{(t+1)}_{\btheta, i} \right] \notag\\
        &\qquad = \left( \E\left[ \accs^{(t+1)}_{\btheta, i} \right] \right)^2 + \Var\left[ \accs^{(t+1)}_{\btheta, i} \right] - \E\left[ \accgamma^{(t+1)}_{\btheta, i} \right] \\
        &\qquad \approx \frac{2 - \beta}{\beta} \left( \E\left[ \Delta \btheta_i^{(t+1)} \right] \right)^2 + \Var\left[ \Delta \btheta_i^{(t+1)} \right] \notag \\
        &\qquad \qquad\qquad - \left( \left( \E\left[ \Delta \btheta_i^{(t+1)} \right] \right)^2 + \Var\left[ \Delta \btheta_i^{(t+1)} \right] \right) \\
        &\qquad = \frac{2 - 2 \beta}{\beta} \left( \E\left[ \Delta \btheta_i^{(t+1)} \right] \right)^2
\end{align}
and 
\begin{align}
        & \E\left[ \accgamma^{(t+1)}_{\btheta, i} \right] - \frac{\beta}{2- \beta}\E\left[ \left( \accs^{(t+1)}_{\btheta, i} \right)^2 \right] \notag \\
        &\qquad \approx \left( \E\left[ \Delta \btheta_i^{(t+1)} \right] \right)^2 + \Var\left[ \Delta \btheta_i^{(t+1)} \right] \notag \\
        &\qquad \qquad - \left( \left( \E\left[ \Delta \btheta_i^{(t+1)} \right] \right)^2 +\frac{\beta}{2 - \beta} \Var\left[ \Delta \btheta_i^{(t+1)} \right] \right) \\
        &\qquad = \frac{2 - 2 \beta}{2 - \beta} \Var\left[ \Delta \btheta_i^{(t+1)} \right] \enspace.
\end{align}
Further, replacing $\E \left[ ( \accs^{(t+1)}_{\btheta, i} )^2 \right]$ and $\E \left[ \accgamma^{(t+1)}_{\btheta, i} \right]$ with $( \accs^{(t+1)}_{\btheta, i} )^2$ and $\accgamma^{(t+1)}_{\btheta, i}$, we obtain the approximations of the squared expectation and the variance of $\Delta \btheta^{(t+1)}_i$ as
\begin{align}
& \left( \E\left[ \Delta \btheta^{(t+1)}_i \right] \right)^2 \approx \frac{\beta}{2 - 2 \beta} \left( \left( \accs^{(t+1)}_{\btheta, i} \right)^2 - \accgamma^{(t+1)}_{\btheta, i} \right) \\
& \Var\left[\Delta \btheta^{(t+1)}_i\right] \approx \frac{2 - \beta}{2 - 2 \beta} \left( \accgamma^{(t+1)}_{\btheta, i} - \frac{\beta \left( \accs^{(t+1)}_{\btheta, i} \right)^2 }{2 - \beta} \right) \enspace.
\end{align}
When $\beta$ is small enough, the variance $\Var\left[\Delta \btheta^{(t+1)}_i\right]$ can be approximated by $\frac{2 - \beta}{2 - 2 \beta} \accgamma^{(t+1)}_{\btheta, i}$ because it holds $\accgamma^{(t+1)}_{\btheta, i} \gg \frac{\beta ( \accs^{(t+1)}_{\btheta, i} )^2 }{2 - \beta}$. Finally, we obtain the approximation of SNR in \eqref{eq:element-snr} as
\begin{align}
\label{eq:element-wise-snr-approx}
\frac{\left(\E\left[\Delta \btheta^{(t+1)}_i\right]\right)^2}{\Var\left[\Delta \btheta^{(t+1)}_i\right]} \approx \frac{\beta}{2 - \beta} \left( \dfrac{\left(\accs_{\btheta, i}^{(t+1)}\right)^2 }{ \accgamma_{\btheta, i}^{(t+1)} } - 1 \right) \enspace.
\end{align}
We set $\beta= 0.01$ as well as in \cite{cmaesled}.

Referred to~\cite{cmaesled}, we combined two estimated element-wise SNRs for $\Delta \bar{\mv}$ and $\Delta \bar{\cov}$ by taking the larger value and ignoring the constant term as 
\begin{align}
        \vsnri^{(t+1)} &= \frac{\beta}{2-\beta} \cdot \max\left( \frac{\left(\accs_{\mv, i}^{(t+1)}\right)^2}{\accgamma_{\mv, i}^{(t+1)}} ,\enspace \frac{\left(\accs_{\cov,i}^{(t+1)}\right)^2}{\accgamma_{\cov,i}^{(t+1)}} \right) \enspace.
\label{eq:compute-vimp}
\end{align}
In addition, $\accs^{(t+1)}_{\btheta, i}$ and $\accgamma^{(t+1)}_{\btheta, i}$ accumulate $\Delta \btheta_i^{(t+1)} / | \Delta \btheta_i^{(t+1)} |$ and $1$ instead of $\Delta \btheta_i^{(t+1)}$ and $( \Delta \btheta_i^{(t+1)} )^2$, respectively (see lines \ref{line:s:m}--\ref{line:gamma:cov} in Algorithm~\ref{alg:proposed}). This modification stabilizes the update in the accumulations.

\begin{algorithm}[t] 
\caption{CMA-ES-LED}
\begin{algorithmic}[1] \label{alg:proposed}
\REQUIRE $\mv^{(0)}, \cov^{(0)}, \sigma^{(0)}$ %\COMMENT{initial distribution parameters}
\REQUIRE $\accs^{(0)}_{\mv} = \accgamma^{(0)}_{\mv} = \accs^{(0)}_{\cov} = \accgamma^{(0)}_{\cov} = \p_v^{(0)} = \mathbf{0}$ %\COMMENT{accumulation parameters}
\REQUIRE $t = 0, \beta=0.01, \lambda = 4 + \lfloor 3 \ln N \rfloor$
\STATE set the hyperparameters $c_c, c_1, c_\mu, c_\sigma$, and $d_\sigma$ as \eqref{eq:hyperparameter}.
\STATE compute $\vthresh$ using \eqref{eq:vthresh}.
\WHILE{termination conditions are not met}
\STATE generate $\lambda$ candidate solutions $\x_1, \cdots, \x_\lambda$. 
\STATE evaluate $\x_1, \cdots, \x_\lambda$ on $f$.
\STATE compute $\Delta \mv^{(t+1)}$ in \eqref{eq:mean-update-direct} and $\Delta_\mu \cov^{(t+1)}$ in \eqref{eq:rank-mu-update}.
\STATE update $\p_c^{(t)}$ by \eqref{eq:rank-one-path-update} and compute $\Delta_1 \boldsymbol{C}^{(t+1)}$ in \eqref{eq:rank-one-update}.
\STATE update $\mv^{(t)}$ and $\cov^{(t)}$ using \eqref{eq:mean-update} and \eqref{eq:cov-update}.
\IF{CSA update}
\STATE update $\p_\sigma^{(t)}$, $\p_v^{(t)}$, and $\sigma^{(t)}$ using \eqref{eq:mod-evolution-path}, \eqref{eq:csa-p-update} and \eqref{eq:csa-led}.
\ELSIF{TPA update}
\STATE update $\sigma^{(t)}$ using \eqref{eq:tpa} and \eqref{eq:tpa-sample-modify}.
\ENDIF
\STATE compute $\Delta \bar{\mv}^{(t+1)}$ in \eqref{eq:bar-mv} and $\Delta \bar{\cov}^{(t+1)}$ in \eqref{eq:bar-cov}.
\FOR {$i = 1$ to $N$}
\STATE $\accs^{(t+1)}_{\mv, i} = (1 - \beta) \accs_{\mv, i}^{(t)} + \sqrt{\beta(2-\beta)} \cdot \frac{\Delta \bar{\mv}_i^{(t+1)}}{| \Delta \bar{\mv}_i^{(t+1)} |}$.
\label{line:s:m}
\STATE $\accgamma^{(t+1)}_{\mv, i} = (1 - \beta)^2 \accgamma_{\mv, i}^{(t)} + \beta(2-\beta) $.
\label{line:gamma:m}
\STATE $\accs^{(t+1)}_{\cov, i} = (1 - \beta) \accs_{\cov, i}^{(t)} + \sqrt{\beta(2-\beta)} \cdot \frac{ \Delta \bar{\cov}_{i}^{(t+1)}}{| \Delta \bar{\cov}_{i}^{(t+1)} |}$.
\label{line:s:cov}
\STATE $\accgamma^{(t+1)}_{\cov, i} = (1 - \beta)^2 \accgamma_{\cov, i}^{(t)} + \beta(2-\beta) $.
\label{line:gamma:cov}
\STATE $\vsnri^{(t+1)} = \frac{\beta}{2-\beta} \cdot \max\left( \frac{\left(\accs_{\mv, i}^{(t+1)}\right)^2}{\accgamma_{\mv, i}^{(t+1)}} ,\enspace \frac{\left(\accs_{\cov,i}^{(t+1)}\right)^2}{\accgamma_{\cov,i}^{(t+1)}} \right)$.
\ENDFOR
\STATE compute $\vgain$ using \eqref{eq:vgain}.
\STATE compute $\vv^{(t+1)}$ using \eqref{eq:compute-vvi} and set $\Neff = \sum_{i=1}^{N} \vv_i^{(t+1)}$.
\STATE adapt the hyperparameters $c_c, c_1, c_\mu, c_\sigma$, and $d_\sigma$ using \eqref{eq:hyperparameter-adaptation} and \eqref{eq:hyperparameter-adaptation-ss} with $\Neff$.
\STATE $t \leftarrow t+1$
\ENDWHILE
\end{algorithmic} 
\end{algorithm}
%
%
%

% ------------------------------
\subsubsection{Transformation of SNR into Effectiveness of Dimension}
% ------------------------------
In this section, we introduce the transformation of $\vsnr^{(t+1)}$ into the estimated effectiveness $\vv^{(t+1)}$.
To explain the required property of such transformation, we compared the dynamics of $\vsnr^{(t)}$ obtained by the CMA-ES with the CSA on the sphere function and the random function that returns a random value as the evaluation value.
We note all dimensions are effective dimensions on the sphere function, and all dimensions are redundant dimensions on the random function.
Figure~\ref{fig:snr} shows the transitions of elements of $\vsnr^{(t)}$ with $N=5$ and $N=100$. We can confirm the transitions of $\vsnr^{(t)}$ on the sphere function and the random function are separable when $N=5$, while they are overlapped when $N=100$. Therefore, we design the transformation of $\vsnr^{(t)}$ to be determined by the search space dimension and sample size of CMA-ES, and the dynamics of $\vsnr^{(t)}$.

We define the effectiveness of each dimension $\vv^{(t)}$ using a monotonically increasing transformation of $\vsnri^{(t+1)}$ as
\begin{align}
        \vv_i^{(t+1)} &= \frac{ \varsigma ( \vsnri^{(t+1)} - \vthresh) }{ \varsigma(1) } \enspace, \label{eq:compute-vvi}
\end{align}
where $\varsigma(x) = 1 / (1 + \exp(- \vgain x))$ is the sigmoid function, and $\vthresh$ and $\vgain$ are parameters adaptively determined. The tuning processes of $\vthresh$ and $\vgain$ are explained as follows.

% ------------------------------
\paragraph{Setting of Threshold Parameter}
% ------------------------------
We tune $\vthresh$ by approximation of the maximum element of $\vsnr^{(t)}$ on the random function. We expect that the elements of $\vv^{(t)}$ on the random function similarly behaves as on the redundant dimensions. We train a regression model of the form 
\begin{align}
\vthresh = \left( a_1 + a_2 \ln N \right) \left( a_3 + a_4 \frac{1}{\sqrt{\lambda}} \right) \enspace. \label{eq:vthresh}
\end{align}
To optimize the coefficients, we run the CMA-ES using the CSA on the random function, varying the number of dimensions and the sample size. All combinations of $N \in \{5n \mid n = 1, \cdots, 100 \}$ and $\lambda \in \{5n \mid n = 1, \cdots, 20 \}$ were performed, and the average values of the maximum of $\vsnr^{(t+1)}$ from 1000-th iteration to 2,000-th iteration were obtained. We performed ten independent trials in each setting. Figure~\ref{fig:maximum-snr} shows the obtained values.

Considering the minimization of the mean squared error between the obtained values and predicted values by \eqref{eq:vthresh}, the coefficients were optimized as $a_1 = 0.106$, $a_2 = 0.0776$, $a_3 = 0.0665$ and $a_4 = 0.947$. The $\mathrm{R}^2$-score of this regression model was $0.9904$.

% ------------------------------
\paragraph{Setting of Gain Parameter}
% ------------------------------
Focusing on the transitions of $\vv^{(t)}$ in the left-side of Figure~\ref{fig:snr}, the transformation~\eqref{eq:compute-vvi} is desired to behave similarly to the step function when $\vsnr^{(t)}$ contains such large elements that are separable from the dynamics on the redundant dimensions. In contrast, when the elements of $\vsnr^{(t)}$ are not separable, as shown in the right-side of Figure~\ref{fig:snr}, the transformation~\eqref{eq:compute-vvi} should always return one to regard all dimensions as effective. As a result, we determine $\vgain$ by a function of the maximum element of $\vsnr$ as
\begin{align}
    \log_{10}\vgain = (g_{\max} - g_{\min})\max(\vsnr) + g_{\min} \enspace. \label{eq:vgain}
\end{align}
We set $g_{\min} = -2$ and $g_{\max} = 3$.

% ------------------------------
\subsection{Improvement of CMA-ES for LED} \label{sec:cmaesled-component}
% ------------------------------
In this section, we introduce two countermeasures for LED using the estimated effectiveness $\vv^{(t+1)}$ and propose CMA-ES-LED. Our countermeasures consist of the hyperparameter adaptation and the refinement of the norm calculation in the step-size adaptation, as explained following.
The update procedure of CMA-ES-LED is summarized in Algorithm~\ref{alg:proposed}.

% ------------------------------
\paragraph{Hyperparameter Adaptation}
% ------------------------------
First, we introduce the hyperparameter adaptation mechanism using $\vv^{(t+1)}$.
We update the hyperparameters using the default values in \eqref{eq:hyperparameter} replacing $N$ with the estimated number of effective dimensions $\Neff = \sum_{i=1}^{N} \vv^{(t+1)}_i$ as
\begin{align}
\begin{aligned}
        c_c &= \frac{4 + \mueff / \Neff}{\Neff + 4 + 2\mueff / \Neff} \enspace  \\
        c_1 &= \frac{2}{(\Neff+1.3)^2 + \mueff} \enspace  \\
        c_\mu &=  \min\left(1-c_1, \frac{2(\mueff-2+1/\mueff)}{(\Neff+2)^2+\mueff}\right) 
\end{aligned} \enspace. \label{eq:hyperparameter-adaptation}
\end{align}
For the step-size update, we set $c_\sigma$ and $d_\sigma$ for the CSA and the TPA as
\begin{align}
\begin{aligned}
        c_\sigma &= \begin{cases}
        \dfrac{\mueff + 2}{\Neff + \mueff + 5} & \hspace{85pt} \text{if CSA} \\
        0.3 & \hspace{85pt} \text{if TPA}
        \end{cases} \\
        d_\sigma &= \begin{cases}
        1 + c_\sigma + 2\max\left(0, \sqrt{\dfrac{\mueff - 1}{\Neff + 1}} - 1\right) & \hspace{4pt} \text{if CSA} \\
        \sqrt{\Neff} & \hspace{4pt} \text{if TPA}
        \end{cases}
\end{aligned} \label{eq:hyperparameter-adaptation-ss}
\end{align}
We note that the sample size is not updated because changing the sample size worsens the estimation accuracy of the SNRs.

\renewcommand{\arraystretch}{1.6}
\begin{table*}[t]
        \caption{List of benchmark functions used in our experiment. Note that $\x_{1:\Ntrueeff} = (\x_1, \cdots, \x_{\Ntrueeff})^\T$ is a $\R^{\Ntrueeff}$ dimensional vector consisting of the first $\Ntrueeff$ elements in $\x$. Before the optimization, we rotated the search space by a random rotation matrix $\Rot$ and obtained the objective function $f: \x \mapsto f_n(\Rot \x)$ for $n=1,\cdots,9$.}
        \vspace{2mm}
        \label{table:benchmark-functions}
        \centering
        \begin{tabular}{cll}
        \hline
        No. & Name & Definition\\
        \hline
        1.&Sphere &  $f_1 (\x) = \sum_{i=1}^{\Ntrueeff} \x_i^2$\\
        2.&Ellipsoid &  $f_2 (\x) = \sum_{i=1}^{\Ntrueeff} 10^{6\frac{i-1}{\Ntrueeff-1}} \x_i^2$\\
        3.&Different Powers & $f_3 (\x) = \sqrt{ \sum^{\Ntrueeff}_{i=1} | \x_i |^{2 + 4 \frac{i-1}{\Ntrueeff-1}}}$\\
        4.&Ackley & $f_4 (\x) = 20 - 20 \exp\left(-0.2 \sqrt{\frac{1}{\Ntrueeff} \sum_{i=1}^{\Ntrueeff} \x_i^2}\right) + \exp(1) - \exp\left(\frac{1}{\Ntrueeff} \sum_{i=1}^{\Ntrueeff}\cos\left(2\pi \x_i \right)\right)$ \\
        5.&Rosenbrock & $f_5 (\x) = \sum_{i=1}^{\Ntrueeff - 1} \left(100(\x_i^2 - \x_{i+1})^2 + (\x_i - 1)^2\right)$\\
        6.&Attractive Sector & $f_6 (\x) = \sum_{i=1}^{\Ntrueeff}(s_i \z_i)^2$, \hspace{5pt} where $\enspace \z_i = 10^{\frac{1}{2}\frac{i-1}{\Ntrueeff-1}} \x_i$, \renewcommand{\arraystretch}{1} $\enspace s_i = \left\{
        \begin{array}{ll}
                10^2 & \text{ if } \z_i > 0\\
                1 & \text{ otherwise }
        \end{array}\right.$\\
        7.&Sharp Ridge & $f_7 (\x) = \x_1^2 + 100\sqrt{\sum_{i=2}^{\Ntrueeff}\x_i^2}$\\
        8.&Bohachevsky & $f_8 (\x) = \sum_{i=1}^{\Ntrueeff-1} \left( \x_i^2 + 2 \x_{i+1}^2 - 0.3 \cos(3 \pi \x_i) - 0.4 \cos(4 \pi \x_{i+1}) + 0.7 \right)$\\
        9.&Rastrigin & $f_9 (\x) = \sum_{i=1}^{\Ntrueeff} \left( \x_i^2 + 10 ( 1- \cos(2 \pi \x_i)) \right)$\\
        \hline
\end{tabular}
\end{table*}
\renewcommand{\arraystretch}{1}
%
%
%

% ------------------------------
\paragraph{Modification of Step-size Adaptations}
% ------------------------------
The redundant dimensions affect the norm calculations in the update rules of the CSA and the TPA, such as the norm of the evolution path $\| \p_\sigma^{(t+1)} \|$ or the norm of the sample form $N$-dimensional Gaussian distribution $\| \N(\0, \I) \|$. This leads to performance degradation on the problem with LED. 
To overcome this issue, we modified the update rules of the CSA and the TPA to measure the norms only on the effective dimensions and to ignore the elements on the redundant dimensions. The refined update rules are described as follows.

% ------------------------------
\paragraph{Modification of CSA}
% ------------------------------
We modified the update rule of evolution path $\p_\sigma^{(t)}$ as 
\begin{align}
        \label{eq:mod-evolution-path}
        \p_\sigma^{(t+1)} = \left(1 - c_\sigma\right)\p_\sigma^{(t)} 
        + \sqrt{c_{\sigma}\left(2-c_{\sigma}\right)\mueff} \sqrt{\vv^{(t)}} \circ \langle \z \rangle_w^{(t+1)},
\end{align}
where $\circ$ is the element-wise product and $\sqrt{\vv} = (\sqrt{\vv_1}, \cdots, \sqrt{\vv_N})^\T$. When $\vv^{(t)}$ stays same point, the law of $\p_\sigma^{(t+1)}$ on the random function is given by a multivariate Gaussian distribution $\N(0, \diag(\p_v^{(t+1)}))$, where $\diag$ returns the diagonal matrix whose diagonal elements are given by the inputted vector, and $\p_v^{(t+1)}$ is the accumulation of $\vv^{(t)}$, i.e., 
\begin{align}
        \p_v^{(t+1)} = (1-c_\sigma)^2\p_v^{(t)} + c_\sigma(2-c_\sigma)\vv^{(t)} \enspace \label{eq:csa-p-update}
\end{align}
with the initial value $\p_v^{(0)} = \0$. While the expected norm of the standard multivariate Gaussian distribution used in the original CSA can be obtained approximately, the calculation of the expected norm of $\N(0, \diag(\p_v^{(t+1)}))$ is intractable. Therefore, we employ the expectation of squared norm analytically derived as
\begin{align}
        p_{v, \mathrm{sum}}^{(t+1)} := \E[ \| \N(0, \diag(\p_v^{(t+1)})) \|^2] = \sum^N_{i=1} \p_{v,i}^{(t+1)} \enspace. 
\end{align}
Then, we obtain the refined update rule of the CSA as
\begin{align}
        \sigma^{(t+1)} = \sigma^{(t)} \exp \left(\frac{c_{\sigma}}{d_{\sigma}}\left(\frac{ \|\p_\sigma^{(t+1)} \|^2}{ p_{v, \mathrm{sum}}^{(t+1)} } - 1\right)\right) \enspace. \label{eq:csa-led}
\end{align}
We also modify the Heaviside function $h^{(t)}$ by replacing $N$ and $\E[\|\N(\mathbf{0}, \I)\|]^2$ with $\Neff$ and $\E[\|\N(\mathbf{0}, \diag(\p_v^{(t+1)}))\|^2]$ in~\eqref{eq:csa-heaviside}.
As a result, we set $h^{(t)} = 1$ when
\begin{align}
        \quad \frac{ \|\p_\sigma^{(t+1)}\|^2 }{ 1 - (1 - c_\sigma )^{2 (t + 1)}} < \left( 1.4 + \frac{2}{\Neff + 1} \right)^2 p_{v, \mathrm{sum}}^{(t+1)} \enspace
\end{align}
and $h^{(t)} = 0$ otherwise.

% ------------------------------
\paragraph{Modification of TPA}
% ------------------------------
Similarly to the modification of the CSA, we modify the generation method of two additional samples in the TPA as 
\begin{align}
        \x_{\pm} = \mv^{(t)} \pm \frac{\sigma^{(t)} \|\N(\0, \I)  \circ \vv^{(t)}\| \cdot \Delta \mv^{(t)}}{ \sqrt{ (\vv^{(t)} \circ \Delta \bar{\mv}^{(t)})^\T (\boldsymbol{\Lambda}^{(t)})^{-1} (\vv^{(t)} \circ \Delta \bar{\mv}^{(t)}) }} \enspace
        \label{eq:tpa-sample-modify}
\end{align}
where $\Delta \bar{\mv}^{(t)}$ is the rotated update direction introduced in \eqref{eq:bar-mv}.
We note that the accumulation~\eqref{eq:tpa-accumulation} and the update rule~\eqref{eq:tpa} are the same as the original TPA.

% ------------------------------
\section{Experiment} \label{sec:experiment}
% ------------------------------

% ------------------------------
\subsection{Experimental Setting} \label{sec:experiment-setting}
% ------------------------------
To demonstrate the performance of CMA-ES-LED on functions with LED, we extended well-known benchmark functions to contain $\Ntrueeff$ effective dimensions and $N - \Ntrueeff$ redundant dimensions. We summarized the benchmark functions in Table~\ref{table:benchmark-functions}. The characteristic of each function is as follows: Sphere $f_1$ is a simple well-conditioned unimodal function. Ellipsoid $f_2$ and Different Powers $f_3$ are ill-conditioned functions. Rosenbrock $f_5$ is non-separable. Attractive Sector $f_6$ is highly asymmetric. Sharp Ridge $f_7$ is non-smooth, non-differentiable, and ill-conditioned. Ackley $f_4$, Bohachevsky $f_8$, and Rastrigin $f_9$ are highly multimodal functions. At the beginning of each trial, we rotated the search space randomly to demonstrate the invariance of CMA-ES-LED to any rotation transformation.

We compared CMA-ES-LED with the original CMA-ES with the CSA and the TPA. The initial mean vector $\mv^{(0)}$ was sampled from $[-5, 5]^N$ uniformly at random. The initial step-size and covariance matrix were set as $\sigma^{(0)} = 2$ and $\cov^{(0)} = \I$, respectively. We regarded a trial as successful when the best evaluation value reached smaller than $10^{-8}$ before the number of evaluations reached $N \times 10^5$. We performed 20 trials for each benchmark function in $f_1, \cdots, f_6$.

In addition, we incorporated CMA-ES-LED into the IPOP restart strategy~\cite{ipopcmaes}, which doubles the sample size and restarts the optimization when any of the stopping criteria is met. We prepared the following stopping criteria.
\begin{itemize}
        \item \texttt{MaxIter}: a trial is terminated if the function evaluations exceeded $100 + 50 (N + 3)^2 / \sqrt{\lambda}$.
        \item \texttt{TolHistFun}: a trial is terminated if the range of the evaluation values of the best sample in each iteration for the last $10 + \lceil 30 N / \lambda \rceil$ iterations was smaller than $10^{-12}$.
        \item \texttt{Stagnation}: we reserved histories of the best and median evaluation values in each iteration over $H_\mathrm{stag}$ iterations, where
        \begin{align*}
                H_\mathrm{stag} = \max\left\{ \min \left\{ 0.2 t, 20000 \right\}, 120 + 30 N / \lambda \right\} \enspace.
        \end{align*}
        Then, the trial was terminated if the medians of the latest $0.3 H_\mathrm{stag}$ values were not better than the medians of the oldest $0.3 H_\mathrm{stag}$ values in both histories.
        \item \texttt{TolX}: a trial is terminated if the square roots of all diagonal components of $(\sigma^{(t)})^2 \cov^{(t)}$ and all components of $\sigma^{(t)} \p_c^{(t)}$ were smaller than $10^{-12} \sigma^{(0)}$.
        \item \texttt{ConditionCov}: a trial is terminated if the condition number of the covariance matrix exceeds $10^{20}$.
\end{itemize}
We selected these stopping criteria from the references~\cite{cmaestutorial, bipop}. For \texttt{ConditionCov}, we increase the upper limit of the condition number from $10^{14}$ to $10^{20}$ because the eigenvalues corresponding to the redundant dimensions will be updated randomly, and it leads to an increase of the condition number easily. We ran 20 trials for each benchmark function in $f_1, \cdots, f_9$. The other experimental setting for the IPOP-CMA-ES is the same as the experimental setting for the CMA-ES.

% ------------------------------
\subsection{Result on Benchmarks with LED} \label{sec:experiment-led}
% ------------------------------
To evaluate CMA-ES-LED on functions with LED, we performed the original CMA-ES and CMA-ES-LED with varying the number of redundant dimensions $N_\mathrm{red} := N - \Ntrueeff$ as $N_\mathrm{red} = 0,4,8,16,32,64,128$, i.e., $N = 8,12,16,24,40,72,136$, fixing the number of effective dimensions as $\Ntrueeff = 8$. 

\begin{figure*}[h!]
\centering
\includegraphics[width=0.75\textwidth]{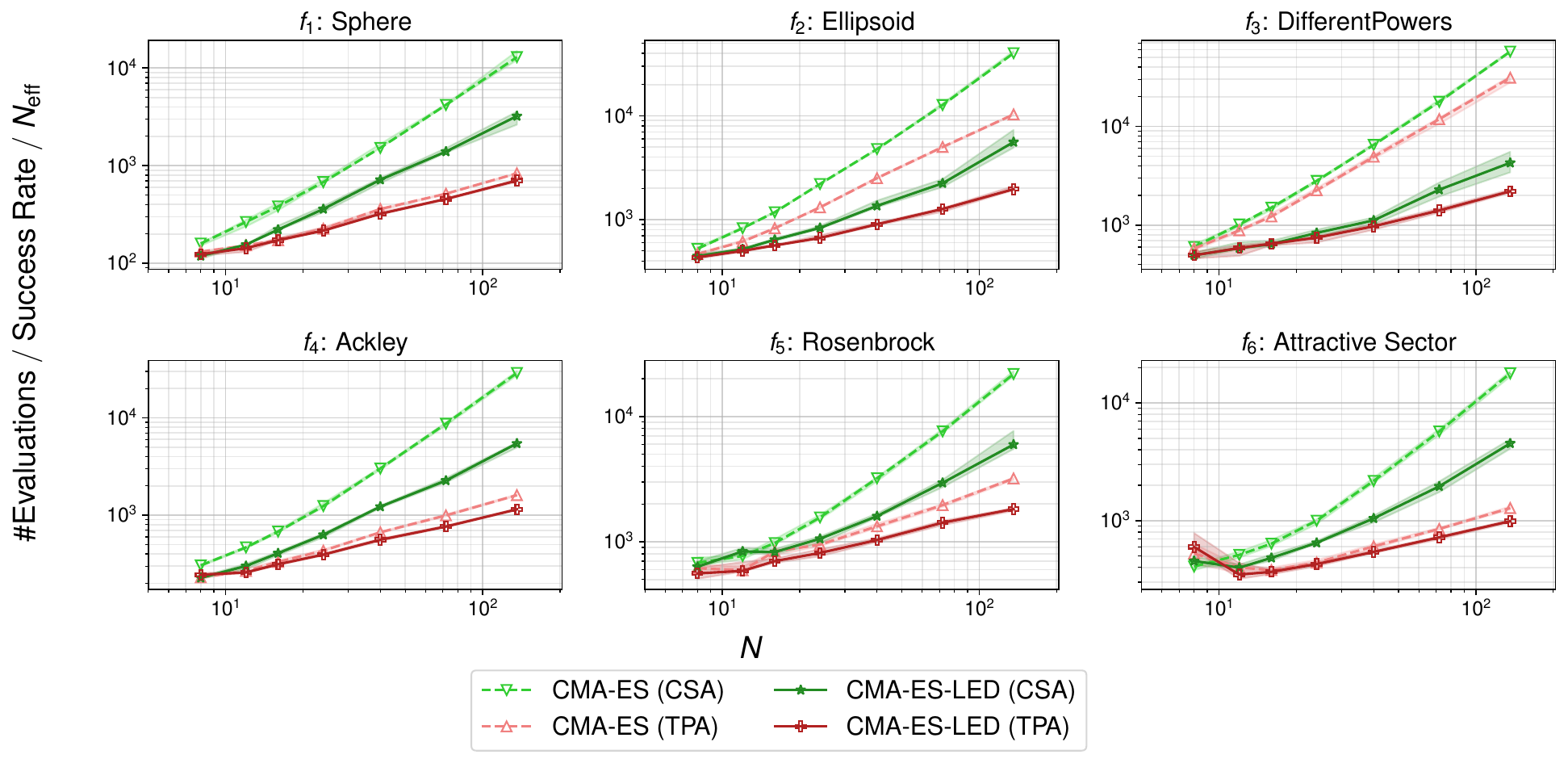}
\caption{
        Comparison of the number of function evaluations divided by the success rate and the number of effective dimensions on the benchmark functions with redundant dimensions.
        The median values and the interquartile ranges over 20 trials are displayed for each $N$.
        The ratio of successful trials is shown when less than 15 trials were successful.
}
\label{fig:add}
\end{figure*}

Figure~\ref{fig:add} depicts the medians and interquartile ranges of the number of function evaluations over the successful trials in the results without the IPOP restart strategy. We also showed the success rate if it is less than $0.75$. 
Regardless of the use of the CSA and TPA, the search performance of CMA-ES-LED was almost the same as the performance of CMA-ES on all functions when $N_\mathrm{red}$ is small, and the performance improvement was gradually increased as $N_\mathrm{red}$ became large. Compared to the case of TPA, more performance improvement was confirmed when using CSA. Significant performance improvements with the TPA were observed on ill-conditioned functions, Ellipsoid $f_2$ and Different Powers $f_3$. We consider that the original update rule of TPA is not significantly affected by the redundant dimension by nature, and such improvement was mainly due to the hyperparameter adaptation, especially the adaptation of learning rates in the covariance matrix update.

\begin{figure*}[h!]
\centering
\includegraphics[width=0.75\textwidth]{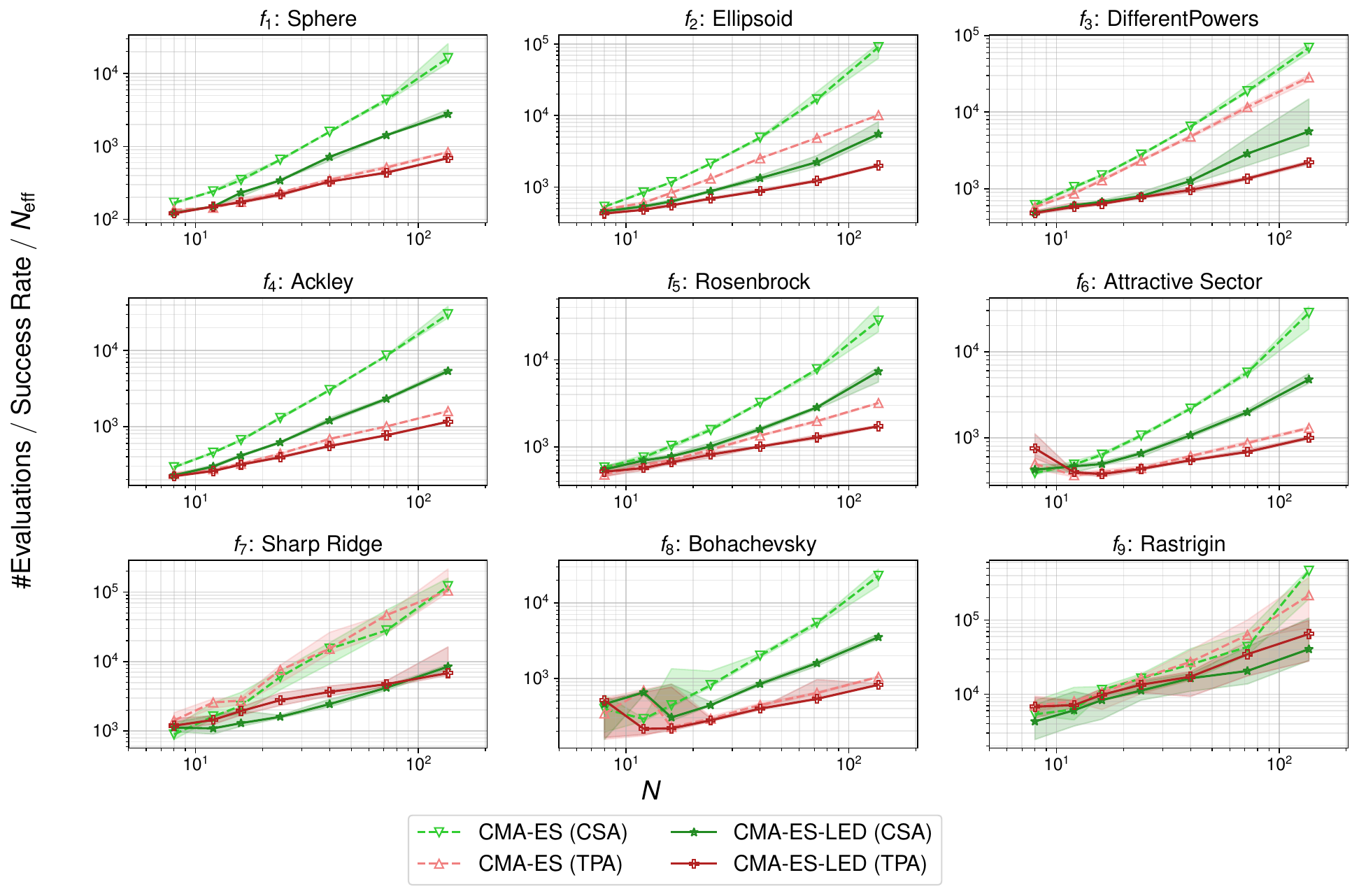}
\caption{
        Comparison of the number of function evaluations divided by the success rate and the number of effective dimensions on the benchmark functions with redundant dimensions. The IPOP restart strategy was applied.
        The median values and the interquartile ranges over 20 trials are displayed for each $N$.
        We note that all trials were successful.
}
\label{fig:add-restart}
\end{figure*}

Figure~\ref{fig:add-restart} shows the results with the IPOP restart strategy. Note that all trials were successful. For the result with the CSA, the search performance was improved on the multimodal functions $f_7$ and $f_8$. However, on Rastrigin $f_9$, the performance improvement was smaller compared with those on other functions. One possible reason for that is that the landscape of Rastrigin makes the estimation of effective dimensions using the estimation of element-wise SNRs unstable. To improve the performance on Rastrigin, other estimation mechanisms of the effective dimensions for highly multimodal functions are required. For the result with the TPA, CMA-ES-LED outperformed the CMA-ES on Sharp Ridge $f_7$. As the Sharp Ridge is ill-conditioned, this may be the effect of the hyperparameter adaptation in the covariance matrix update, as well on Ellipsoid $f_2$ and Different Powers $f_3$.

\begin{figure*}[h!]
\centering
\includegraphics[width=0.75\textwidth]{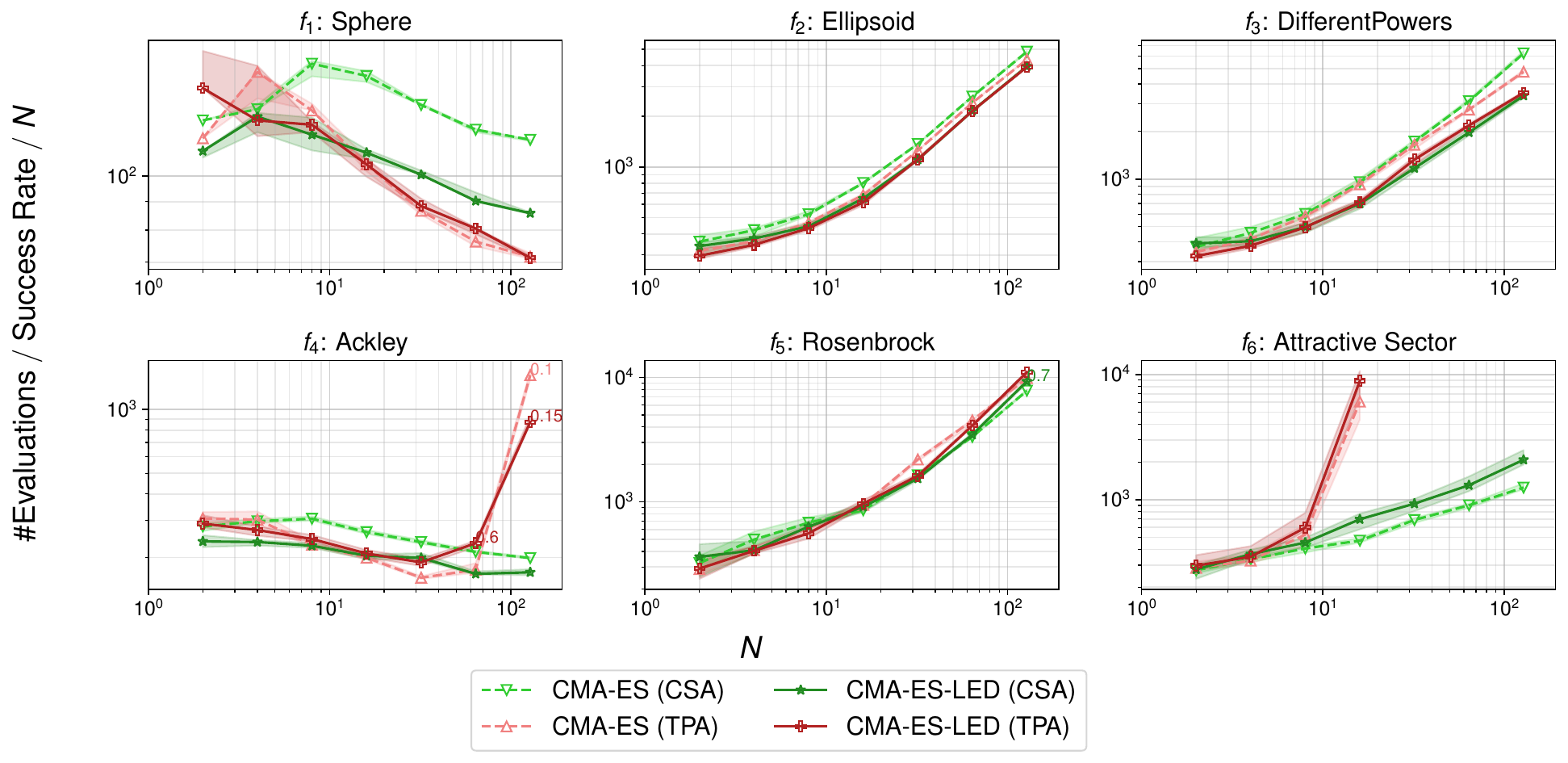}
\caption{
        Comparison of the number of function evaluations divided by the success rate and the number of dimensions on the benchmark functions without redundant dimensions.
        The median values and the interquartile ranges over 20 trials are displayed for each $N$.
        The ratio of successful trials is shown when less than 15 trials were successful.
}
\label{fig:normal}
\end{figure*}

\begin{figure*}[h!]
\centering
\includegraphics[width=0.75\textwidth]{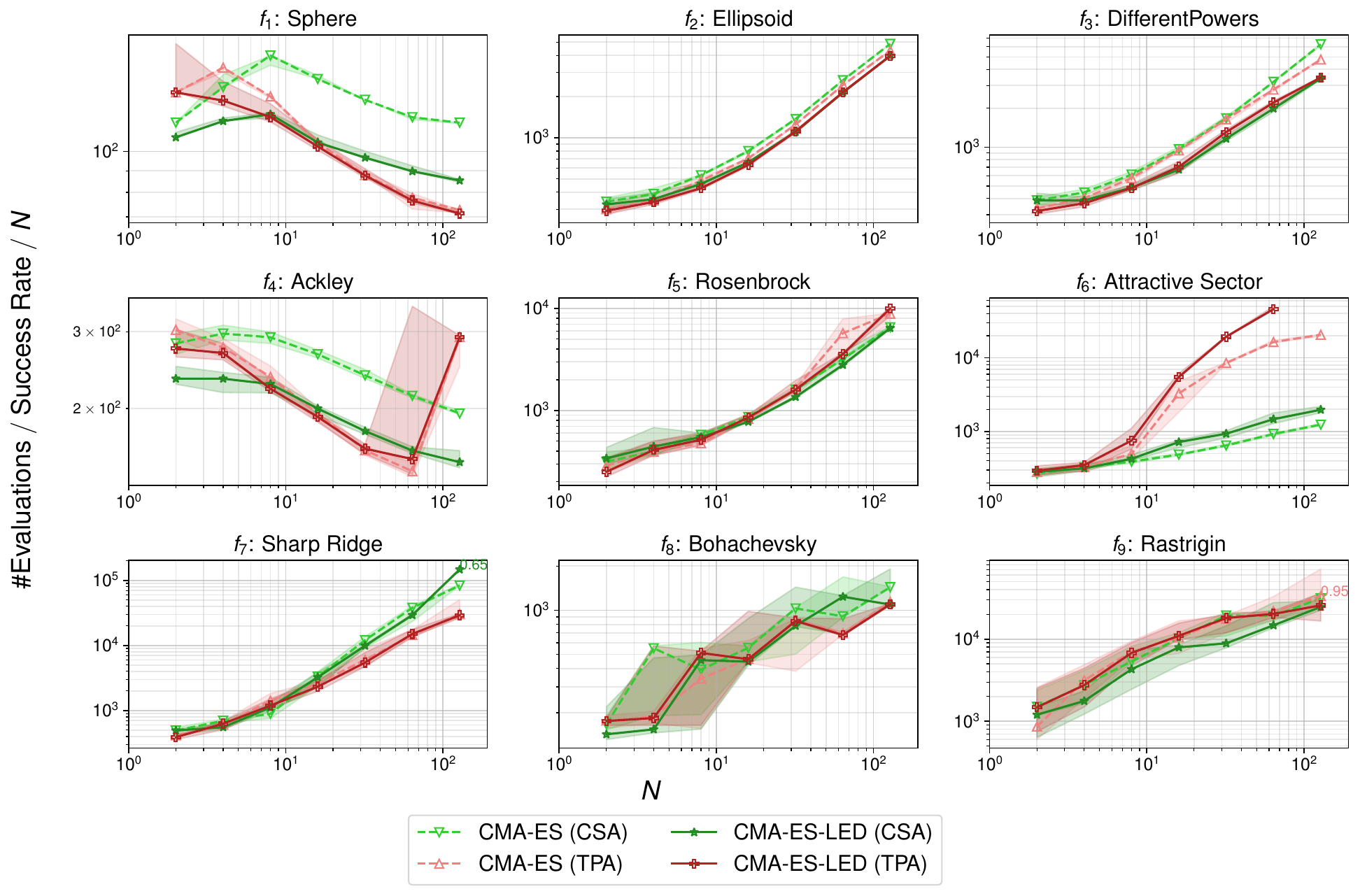}
\caption{
        Comparison of the number of function evaluations divided by the success rate and the number of dimensions on the benchmark functions without redundant dimensions.
        The IPOP restart strategy was applied.
        The median values and the interquartile ranges over 20 trials are displayed for each $N$.
        The ratio of successful trials is shown when unsuccessful trial exist.
}
\label{fig:normal-restart}
\end{figure*}
%
%
%

% ------------------------------
\subsection{Result on Benchmarks without LED} \label{sec:experiment-normal}
% ------------------------------
We show the experimental result on the benchmark functions without LED, i.e., $N = \Ntrueeff$. We performed trials changing the total number of dimensions as $N = 2,4,8, \cdots, 128$.
We note that the CMA-ES-LED is designed to improve the search performance on functions with LED, as described in Section~\ref{sec:experiment-led}. Therefore, it is acceptable if no performance improvements were observed on the functions without LED.

Figure~\ref{fig:normal} shows the medians and interquartile ranges of the number of function evaluations over the successful trials. Figure~\ref{fig:normal} shows the result without the IPOP restart strategy and denotes the success rate if it is less than $0.75$. Focusing the case with CSA, the search performance was slightly improved by our method on high-dimensional Sphere $f_1$, Different Powers $f_3$, and Ackley $f_4$. We consider two reasons for this improvement. The first reason is that the bias in the estimated effectiveness $\vv^{(t)}$ of each dimension increased the learning rates and accelerated the optimization. The second reason is that, due to the modification of the CSA, the elements of evolution path $\p_\sigma^{(t)}$ corresponding to high SNRs were enhanced, and the step-size was updated profoundly. In contrast, the performance on Attractive Sector $f_6$ was worsened slightly. However, serious performance deterioration was not confirmed.
Focusing on the case of the TPA, the performance of CMA-ES-LED is almost the same as the original CMA-ES. 
In contrast to the case of CSA, there were no performance improvements on Sphere $f_1$ and Ackley $f_4$. As the modifications of the TPA, we modified the generation process of two additional solutions only, which is considered to have less effect than the modifications in the CSA. 

Figure~\ref{fig:normal-restart} shows the result with the IPOP restart strategy. We denote the success rate if there was at least one unsuccessful trial. 
The case of the IPOP restart strategy shows similar tendencies observed in the case of no restart strategy.
In addition, any severe performance degradation was not confirmed on multimodal functions $f_4, f_8, f_9$. This showed the robustness of our estimation process of the effective dimensions on the multimodal landscape.
Focusing on Attractive Sector $f_6$, both the CMA-ES and CMA-ES-LED using the TPA could optimize it successfully while they failed without IPOP restart strategy. Moreover, the CMA-ES-LED was worse than CMA-ES on $f_6$ in both cases where the CSA and TPA were used. We consider the reason is that the performance of the CMA-ES on Attractive Sector is sensitive to the hyperparameter setting, and our hyperparameter adaptation mechanism leads to an unsuitable hyperparameter setting.

\begin{figure*}[h!]
\centering
\includegraphics[width=0.78\textwidth]{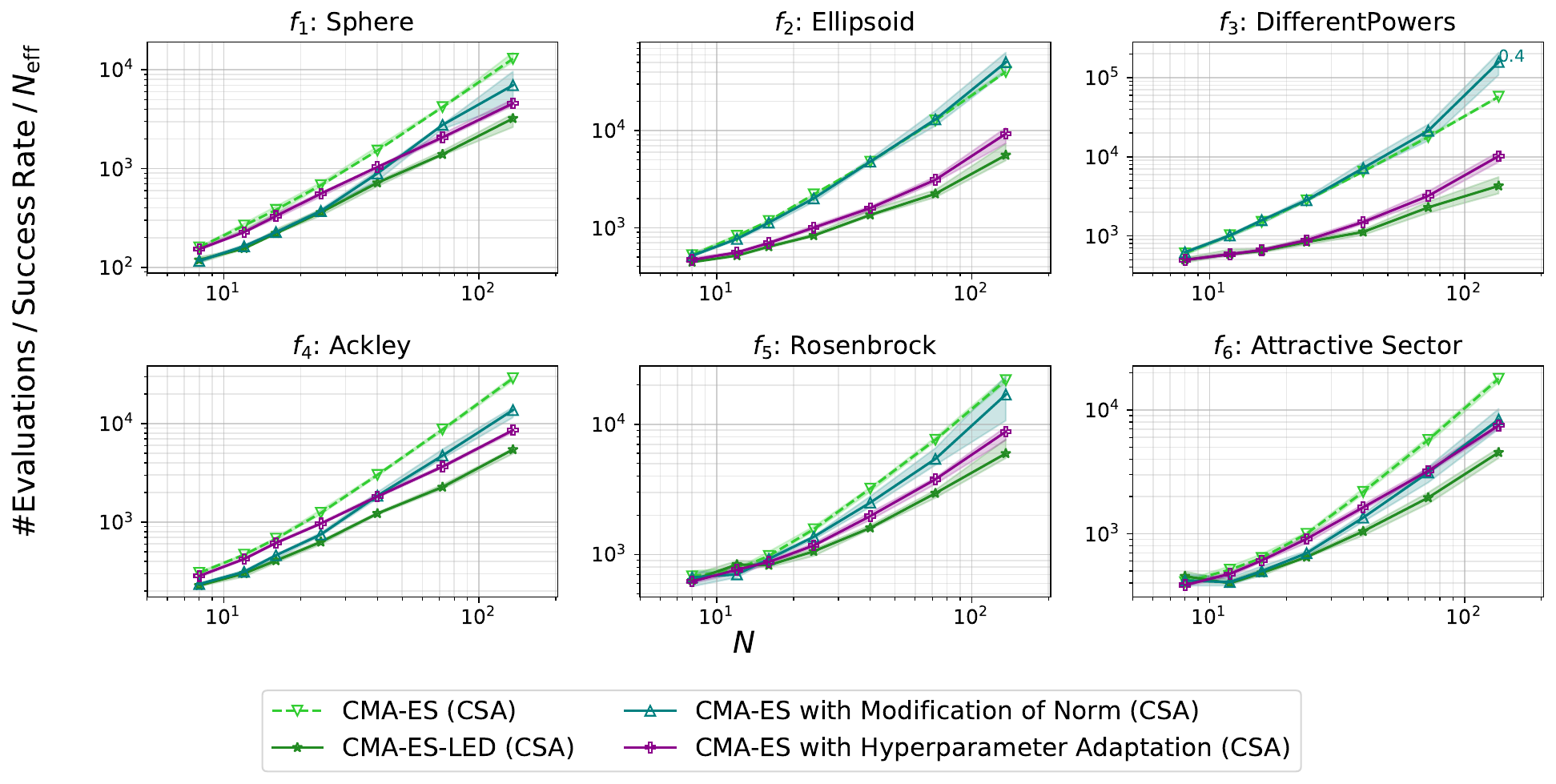}
\caption{
        Comparison of ablations with CSA on the benchmark functions with redundant dimensions. We plot the number of function evaluations divided by the success rate and the number of effective dimensions.
        The median values and the interquartile ranges over 20 trials are displayed for each $N$.
        The ratio of successful trials is shown when less than 15 trials were successful.
}
\label{fig:abl-csa}
\end{figure*}

\begin{figure*}[h!]
\centering
\includegraphics[width=0.78\textwidth]{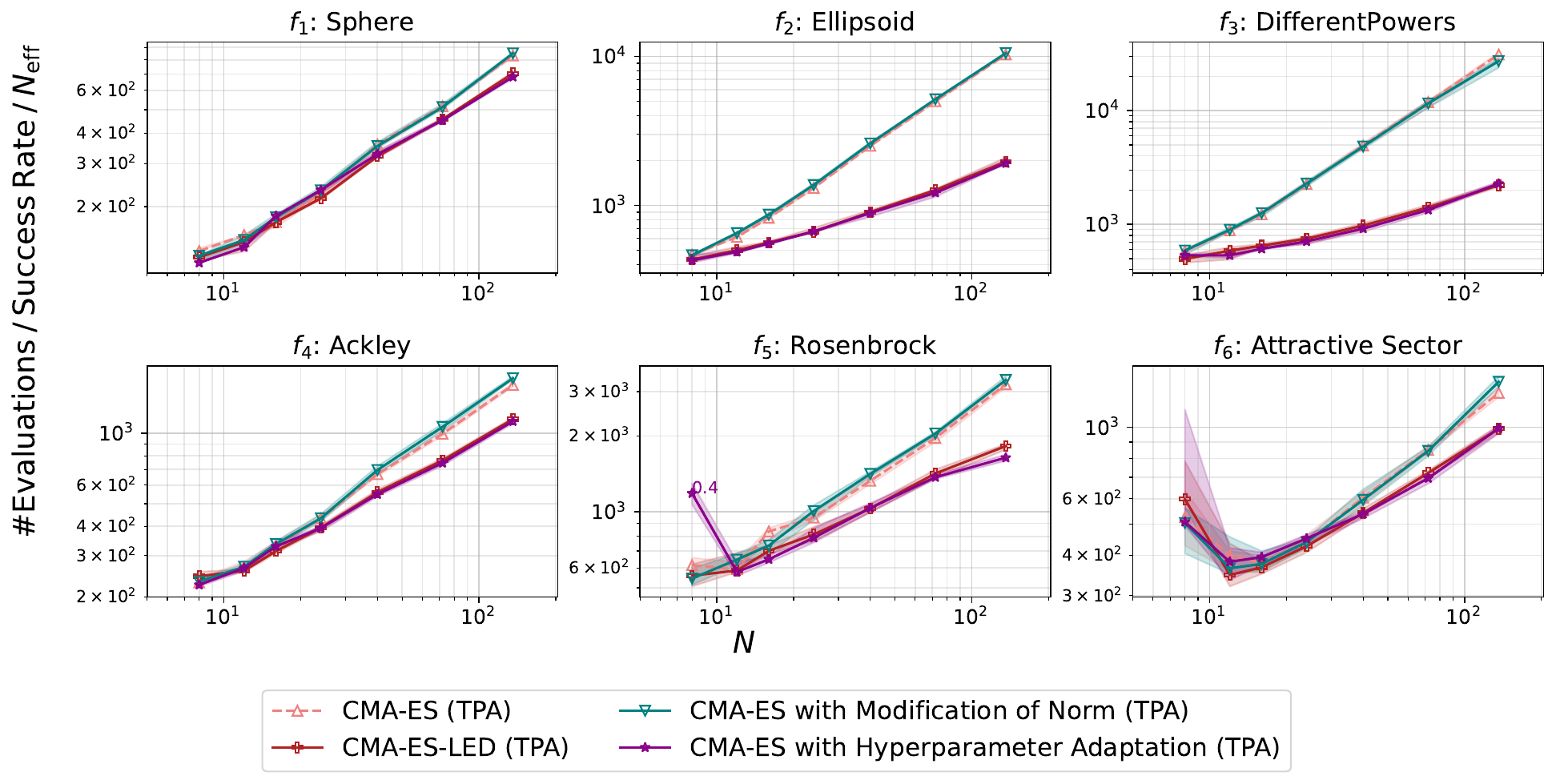}
\caption{
        Comparison of ablations with TPA on the benchmark functions with redundant dimensions. We plot the number of function evaluations divided by the success rate and the number of effective dimensions. 
        The median values and the interquartile ranges over 20 trials are displayed for each $N$.
        The ratio of successful trials is shown when less than 15 trials were successful.
}
\label{fig:abl-tpa}
\end{figure*}
%
%
%

% ------------------------------
\subsection{Result of Ablation Study} \label{sec:experiment-abb}
% ------------------------------
We evaluated each mechanism of the CMA-ES-LED by the ablation study. We performed two ablations of CMA-ES-LED, the CMA-ES with our hyperparameter adaptation and the CMA-ES with the modification of norm calculation in the step-size adaptation.
The experimental setting was the same as in Section~\ref{sec:experiment-led}.
Figure~\ref{fig:abl-csa} and Figure~\ref{fig:abl-tpa} show the result using the CSA and TPA, respectively. When comparing the CMA-ES with the modification of norm calculation to the original CMA-ES, the modification of norm calculation contributed to performance improvement when using CSA, while it did not when using the TPA. On the other hand, the hyperparameter adaptation works efficiently in both cases. This implies the combination of our hyperparameter adaptation with other step-size adaptations which do not use norm calculation, such as the median success rule~\cite{msr}, is also a promising approach to tackle the LED property. Moreover, for the result with the CSA on ill-conditioned functions $f_2$ and $f_3$, the hyperparameter adaptation works well rather than the modification of norm calculation, as well as the result with the TPA. We consider that the hyperparameter setting is sensitive for ill-conditioned functions, which implies the importance of the hyperparameter adaptation mechanism not only for functions with LED.

% ------------------------------
\section{Conclusion} \label{sec:conclusion}
% ------------------------------
This study proposed the CMA-ES-LED, an improved variant of the CMA-ES, to tackle the functions with LED. To reconstruct the intrinsic objective function from the objective function, we estimate the effectiveness of each dimension in the rotated search space. The rotation matrix is obtained by the eigenvectors of the covariance matrix. We also introduce a monotonically increasing function to obtain the estimated effectiveness of each dimension based on the estimated element-wise SNRs of the update directions of the mean vector and the rank-$\mu$ update. The parameters of the function are adaptively determined without additional parameter tuning by the user. Then, we proposed two countermeasures for LED, 1) the hyperparameter adaptation based on the estimated number of effective dimensions, and 2) the refinement of the norm calculation in the CSA and the TPA to measure it only on the effective dimensions. We confirmed the improvement of CMA-ES-LED over the original CMA-ES on the benchmark functions with LED, including the cases where the IPOP restart strategy was incorporated. We also confirmed that the CMA-ES-LED did not deteriorate the search performance on functions without LED.

In the restart strategy, the estimated effectiveness of each dimension and the rotation matrix are initialized since the covariance matrix is also initialized. The development of a mechanism to inherit the rotation matrix at restarting may improve the performance, which is left as future work. In addition, as we fixed the sample size to the default setting, combining the population size adaptation~\cite{psacmaes} is another interesting future work.

% ------------------------------
\section*{Acknowledgement} 
% ------------------------------
This work was partially supported by JSPS KAKENHI (JP20H04240, 23H03466, 24K20857), JST PRESTO (JPMJPR2133), ACT-X (JPMJAX24C7), and a project commissioned by NEDO (JPNP18002, JPNP20006).

\bibliographystyle{elsarticle-num} 
\bibliography{bibliography}

\begin{thebibliography}{10}
\expandafter\ifx\csname url\endcsname\relax
  \def\url#1{\texttt{#1}}\fi
\expandafter\ifx\csname urlprefix\endcsname\relax\def\urlprefix{URL }\fi
\expandafter\ifx\csname href\endcsname\relax
  \def\href#1#2{#2} \def\path#1{#1}\fi

\bibitem{bboapplication}
I.~Bajaj, A.~Arora, M.~M.~F. Hasan, Black-Box Optimization: Methods and
  Applications, Springer International Publishing, Cham, 2021, pp. 35--65.
\newblock \href {https://doi.org/10.1007/978-3-030-66515-9\_2}
  {\path{doi:10.1007/978-3-030-66515-9\_2}}.

\bibitem{cmaes}
N.~Hansen, S.~D. M{\"u}ller, P.~Koumoutsakos, Reducing the time complexity of
  the derandomized evolution strategy with covariance matrix adaptation
  ({CMA-ES}), Evol. Comput. 11~(1) (2003) 1--18.

\bibitem{cmaestutorial}
N.~Hansen, The {CMA} evolution strategy: {A} tutorial, CoRRArXiv:1604.00772
  (2016).

\bibitem{led}
R.~E. Caflisch, W.~Morokoff, A.~Owen, Valuation of mortgage-backed securities
  using brownian bridges to reduce effective dimension, J. Comput. Finance
  1~(1) (1997) 27--46.
\newblock \href {https://doi.org/10.21314/JCF.1997.005}
  {\path{doi:10.21314/JCF.1997.005}}.

\bibitem{mlhyperopt}
J.~Bergstra, Y.~Bengio, Random search for hyper-parameter optimization, J.
  Mach. Learn. Res. 13 (2012) 281--305.

\bibitem{led-example-airplane-control}
A.~Khelassi, P.~Weber, D.~Theilliol, Reconfigurable control design for
  over-actuated systems based on reliability indicators, in: 2010 Conference on
  Control and Fault-Tolerant Systems (SysTol), 2010, pp. 365--370.
\newblock \href {https://doi.org/10.1109/SYSTOL.2010.5675957}
  {\path{doi:10.1109/SYSTOL.2010.5675957}}.

\bibitem{led-example-aircraft-wing-design}
T.~Lukaczyk, P.~Constantine, F.~Palacios, J.~Alonso, Active subspaces for shape
  optimization, in: Proceedings of the 10th AIAA Multidisciplinary Design
  Optimization Conference, 2014, pp. 1--18.

\bibitem{csa1}
N.~Hansen, A.~Ostermeier, Completely derandomized self-adaptation in evolution
  strategies, Evol. Comput. 9~(2) (2001) 159--195.

\bibitem{tpa1}
N.~Hansen, A.~Atamna, A.~Auger, How to assess step-size adaptation mechanisms
  in randomised search, in: Proceedings of Parallel Problem Solving from Nature
  (PPSN), Vol. 8672 of Lecture Notes in Computer Science, Springer, 2014, pp.
  60--69.

\bibitem{rembo}
Z.~Wang, F.~Hutter, M.~Zoghi, D.~Matheson, N.~de~Freitas, Bayesian optimization
  in a billion dimensions via random embeddings, J. Artif. Intell. Res. 55
  (2016) 361--387.

\bibitem{remeda}
M.~L. Sanyang, A.~Kab{\.a}n, {REMEDA}: {R}andom embedding {EDA} for optimising
  functions with intrinsic dimension, in: Proceedings of Parallel Problem
  Solving from Nature (PPSN), Vol. 9921 of Lecture Notes in Computer Science,
  Springer, 2016, pp. 859--868.

\bibitem{ReMO:2017}
H.~Qian, Y.~Yu, Solving high-dimensional multi-objective optimization problems
  with low effective dimensions, in: Proceedings of the AAAI Conference on
  Artificial Intelligence, Vol.~31, 2017.
\newblock \href {https://doi.org/10.1609/aaai.v31i1.10664}
  {\path{doi:10.1609/aaai.v31i1.10664}}.

\bibitem{REGO:2022}
C.~Cartis, E.~Massart, A.~Otemissov, Global optimization using random
  embeddings, Math. Program. (2022).
\newblock \href {https://doi.org/10.1007/s10107-022-01871-y}
  {\path{doi:10.1007/s10107-022-01871-y}}.

\bibitem{boundREGO:2023}
C.~Cartis, E.~Massart, A.~Otemissov,
  \href{https://doi.org/10.1007/s10107-022-01812-9}{Bound-constrained global
  optimization of functions with low effective dimensionality using multiple
  random embeddings}, Math. Program. 198~(1) (2023) 997--1058.
\newblock \href {https://doi.org/10.1007/s10107-022-01812-9}
  {\path{doi:10.1007/s10107-022-01812-9}}.
\newline\urlprefix\url{https://doi.org/10.1007/s10107-022-01812-9}

\bibitem{asngled}
T.~Yamaguchi, K.~Uchida, S.~Shirakawa, Adaptive stochastic natural gradient
  method for optimizing functions with low effective dimensionality, in:
  Proceedings of Parallel Problem Solving from Nature (PPSN), Vol. 12269 of
  Lecture Notes in Computer Science, Springer, 2020, pp. 719--731.

\bibitem{asng}
Y.~Akimoto, S.~Shirakawa, N.~Yoshinari, K.~Uchida, S.~Saito, K.~Nishida,
  Adaptive stochastic natural gradient method for one-shot neural architecture
  search, in: International Conference on Machine Learning (ICML), Vol.~97 of
  Proceedings of Machine Learning Research, PMLR, 2019, pp. 171--180.

\bibitem{bidirectional}
Y.~Akimoto, Y.~Nagata, I.~Ono, S.~Kobayashi, Bidirectional relation between cma
  evolution strategies and natural evolution strategies, in: Proceedings of
  Parallel Problem Solving from Nature (PPSN), Vol. 6238 of Lecture Notes in
  Computer Science, Springer, 2010, pp. 154--163.

\bibitem{ipopcmaes}
A.~Auger, N.~Hansen, A restart {CMA} evolution strategy with increasing
  population size, in: 2005 IEEE Congress on Evolutionary Computation, Vol.~2,
  2005, pp. 1769--1776.
\newblock \href {https://doi.org/10.1109/CEC.2005.1554902}
  {\path{doi:10.1109/CEC.2005.1554902}}.

\bibitem{cmaesled}
T.~Yamaguchi, K.~Uchida, S.~Shirakawa, Improvement of sep-{CMA-ES} for
  optimization of high-dimensional functions with low effective dimensionality,
  in: 2022 IEEE Symposium Series on Computational Intelligence (SSCI), 2022,
  pp. 1659--1668.
\newblock \href {https://doi.org/10.1109/SSCI51031.2022.10022244}
  {\path{doi:10.1109/SSCI51031.2022.10022244}}.

\bibitem{sepcmaes}
R.~Ros, N.~Hansen, A simple modification in {CMA-ES} achieving linear time and
  space complexity, in: Proceedings of Parallel Problem Solving from Nature
  (PPSN), Vol. 5199 of Lecture Notes in Computer Science, Springer, 2008, pp.
  296--305.

\bibitem{vkdcmaes}
Y.~Akimoto, N.~Hansen, Projection-based restricted covariance matrix adaptation
  for high dimension, in: Proceedings of the Genetic and Evolutionary
  Computation Conference (GECCO), ACM, 2016, pp. 197--204.
\newblock \href {https://doi.org/10.1145/2908812.2908863}
  {\path{doi:10.1145/2908812.2908863}}.

\bibitem{igo}
Y.~Ollivier, L.~Arnold, A.~Auger, N.~Hansen, Information-geometric optimization
  algorithms: {A} unifying picture via invariance principles, J. Mach. Learn.
  Res. 18~(18) (2017) 1--65.

\bibitem{naturalgradient}
S.~Amari, Natural gradient works efficiently in learning, Neural Comput. 10~(2)
  (1998) 251--276.

\bibitem{bipop}
N.~Hansen, Benchmarking a bi-population {CMA-ES} on the {BBOB-2009} function
  testbed, in: Proceedings of the 11th Annual Conference Companion on Genetic
  and Evolutionary Computation Conference: Late Breaking Papers, 2009, pp.
  2389--2396.
\newblock \href {https://doi.org/10.1145/1570256.1570333}
  {\path{doi:10.1145/1570256.1570333}}.

\bibitem{msr}
O.~A. ElHara, A.~Auger, N.~Hansen, A median success rule for non-elitist
  evolution strategies: {S}tudy of feasibility, in: Proceedings of the
  Conference on Genetic and Evolutionary Computation (GECCO), ACM, 2013, pp.
  415--422.

\bibitem{psacmaes}
K.~Nishida, Y.~Akimoto, {PSA-CMA-ES}: {CMA-ES} with population size adaptation,
  in: Proceedings of the Genetic and Evolutionary Computation Conference
  (GECCO), ACM, 2018, pp. 865--872.
\newblock \href {https://doi.org/10.1145/3205455.3205467}
  {\path{doi:10.1145/3205455.3205467}}.

\end{thebibliography}

\end{document}